\documentclass[sigconf]{acmart}
\AtBeginDocument{%
  \providecommand\BibTeX{{%
    \normalfont B\kern-0.5em{\scshape i\kern-0.25em b}\kern-0.8em\TeX}}}

\copyrightyear{2024}
\acmYear{2024}
\setcopyright{acmlicensed}
\acmConference[KDD '24]{Proceedings of the 30th ACM SIGKDD Conference on Knowledge Discovery and Data Mining}{August 25--29, 2024}{Barcelona, Spain}
\acmBooktitle{Proceedings of the 30th ACM SIGKDD Conference on Knowledge Discovery and Data Mining (KDD '24), August 25--29, 2024, Barcelona, Spain}
\acmDOI{10.1145/3637528.3671760}
\acmISBN{979-8-4007-0490-1/24/08}

\usepackage{hyperref}
\hypersetup{
    colorlinks=true,
    linkcolor=black,
    urlcolor=black,
    citecolor=black
}

\def\equationautorefname~#1\null{Equation~(#1)\null}
\usepackage{url}
\usepackage{bm}
\usepackage{bbm}
\usepackage{xspace}
\usepackage{algorithmic}
\usepackage{algorithm}
\usepackage{mdwlist}    
\usepackage{paralist} 

\usepackage{adjustbox}
\usepackage{array}
\usepackage{booktabs}
\usepackage{multirow}

\usepackage{array,graphicx}
\usepackage{booktabs}
\usepackage{pifont}
\usepackage{color}

\usepackage{colortbl}
\definecolor{lightgray}{gray}{0.85}
\usepackage{multirow}

\usepackage{fancybox} 
\usepackage{amsmath}
\usepackage{amsfonts}

\usepackage{enumitem}

\newcommand{\alg}[1]{Alg.~#1}
\newcommand{\step}[1]{Line~#1}
\newcommand{\eq}[1]{Eq.~(#1)}
\newcommand{\tabl}[1]{Table~#1}
\newcommand{\fig}[1]{Fig.~#1}
\newcommand{\secton}[1]{Section~#1}
\newcommand{\apdx}[1]{Appendix~#1}

\newcommand{\myparaitemize}[1]{\noindent{\textbf{#1.}}}

\newtheorem{problem}{Problem}
\newtheorem{lemma}{Lemma}

\newcommand{\romanone}{$\mathrm{(\hspace{.18em}i\hspace{.18em})}$\xspace}
\newcommand{\romantwo}{$\mathrm{(\hspace{.08em}ii\hspace{.08em})}$\xspace}
\newcommand{\romanthree}{$\mathrm{(i\hspace{-.08em}i\hspace{-.08em}i)}$\xspace}

\newcommand{\hide}[1]{}

\newcommand{\method}{\textsc{MissNet}\xspace}

\newcommand{\thN}[1]{$#1^{th}$\xspace}

\newcommand{\lone}{$\ell_1$\xspace}
\newcommand{\lonenorm}{\lone-norm\xspace}

\newcommand{\mts}{\bm{X}} 
\newcommand{\indicatormatrix}{\bm{W}} 
\newcommand{\mtsvector}{\bm{x}} 

\newcommand{\givenmts}{\tilde{\mts}} 
\newcommand{\givenmtsvector}{\tilde{\mtsvector}} 

\newcommand{\impmts}{\hat{\mts}} 
\newcommand{\impmtsvector}{\hat{\mtsvector}} 

\newcommand{\timestep}{T} 
\newcommand{\ndim}{N} 
\newcommand{\latentdim}{L} 

\newcommand{\latentfactor}{\bm{Z}} 
\newcommand{\objectfactor}{\bm{V}} 
\newcommand{\objectfactorN}[1]{\objectfactor^{(#1)}} 
\newcommand{\objectfactorset}{\dot{\objectfactor}} 
\newcommand{\switchmatrix}{\bm{F}} 
\newcommand{\latentfactorvector}{\bm{z}} 
\newcommand{\objectfactorvector}{\bm{v}} 
\newcommand{\objectfactorvectorN}[1]{\objectfactorvector^{(#1)}} 
\newcommand{\switchvector}{\bm{f}} 

\newcommand{\parameter}{\theta} 
\newcommand{\contextualmatrix}{\bm{S}} 
\newcommand{\contextualmatrixN}[1]{\contextualmatrix^{(#1)}}
\newcommand{\contextualmatrixset}{\dot{\contextualmatrix}} 
\newcommand{\objectmatrix}{\bm{U}} 
\newcommand{\objectmatrixN}[1]{\objectmatrix^{(#1)}}
\newcommand{\objectmatrixset}{\dot{\objectmatrix}} 
\newcommand{\transitionmatrix}{\bm{B}} 

\newcommand{\meanvector}{\bm{\rho}} 
\newcommand{\meanvectorN}[1]{\meanvector^{(#1)}}
\newcommand{\meanvectorset}{\dot{\meanvector}} 
\newcommand{\invcovmatrix}{\bm{\Theta}} 
\newcommand{\invcovmatrixN}[1]{\invcovmatrix^{(#1)}}
\newcommand{\invcovmatrixset}{\dot{\invcovmatrix}} 
\newcommand{\contextualvector}{\bm{s}} 
\newcommand{\contextualvectorN}[1]{\contextualvector^{(#1)}}

\newcommand{\sgmZero}{\bm{\Psi}_{0}} 
\newcommand{\sgm}{\sigma} 
\newcommand{\sgmZ}{\sgm_{\latentfactor}} 

\newcommand{\sgmX}{\sgm_{\mts}} 
\newcommand{\sgmXN}[1]{\sgm_{\mts^{(#1)}}}
\newcommand{\sgmXset}{\dot{\sgmX}}
\newcommand{\sgmS}{\sgm_{\contextualmatrix}} 
\newcommand{\sgmSN}[1]{\sgm_{\contextualmatrix^{(#1)}}}
\newcommand{\sgmSset}{\dot{\sgmS}}
\newcommand{\sgmV}{\sgm_{\objectfactor}} 
\newcommand{\sgmVN}[1]{\sgm_{\objectfactor^{(#1)}}}
\newcommand{\sgmVset}{\dot{\sgmV}}

\newcommand{\Sgm}{\bm{\Sigma}} 
\newcommand{\SgmZ}{\Sgm_{\latentfactor}} 

\newcommand{\SgmXN}[1]{\Sgm_{\mts^{(#1)}}}

\newcommand{\SgmSN}[1]{\Sgm_{\contextualmatrix^{(#1)}}}

\newcommand{\SgmVN}[1]{\Sgm_{\objectfactor^{(#1)}}}

\newcommand{\mcini}{\bm{\pi}_0} 
\newcommand{\mctra}{\bm{\Pi}} 

\newcommand{\nstate}{K} 
\newcommand{\contextparam}{\alpha} 
\newcommand{\sparseparam}{\lambda} 

\newcommand{\constant}{e} 

\newcommand{\setK}[1]{\{ #1 \}_{k=1}^{\nstate}} 

\newcommand{\kalmangain}{\bm{D}} 
\newcommand{\formean}{\bm{\mu}} 
\newcommand{\forcov}{\bm{\Psi}} 

\newcommand{\kalmanQ}{\bm{C}} 
\newcommand{\bacmean}{\bm{\hat{\mu}}} 
\newcommand{\baccov}{\bm{\hat{\Psi}}} 

\newcommand{\bayM}{\bm{M}}
\newcommand{\baymean}{\bm{\nu}}
\newcommand{\baycov}{\bm{\Upsilon}}
\newcommand{\bayMN}[1]{\bayM^{(#1)}}
\newcommand{\baymeanN}[1]{\baymean^{(#1)}}
\newcommand{\baycovN}[1]{\baycov^{(#1)}}

\newcommand{\Uone}{\bm{A_{1}}}
\newcommand{\UoneN}[1]{\Uone^{(#1)}}
\newcommand{\Utwo}{\bm{A_{2}}}
\newcommand{\UtwoN}[1]{\Utwo^{(#1)}}

\newcommand{\obsindex}{\bm{o}} 
\newcommand{\obsmtsvector}{\bar{\mtsvector}} 
\newcommand{\obsobjectmatrixN}[1]{\bar{\objectmatrix}^{(#1)}}
\newcommand{\obsobjectmatrixNinv}[1]{\bar{\objectmatrix}^{(#1)'}}

\newcommand{\switchcost}{J} 
\newcommand{\invsig}{\bm{R}} 

\newcommand{\idmat}{\bm{I}} 

\newcommand{\realnumber}{\mathbb{R}} 

\newcommand{\gauss}[2]{\mathcal{N}(#1, #2)} 
\newcommand{\probability}[1]{p(#1)} 
\newcommand{\conditionprobability}[2]{p(#1|#2)}
\newcommand{\expectation}[1]{\mathbb{E}[#1]}
\newcommand{\expectationunder}[2]{\mathbb{E}_{#2}[#1]}
\newcommand{\objective}[1]{Q(#1)} 

\newcommand{\loglike}{ll}

\newcommand{\hadamard}{\circ} 

\newcommand{\argmax}{\mathop{\rm arg~max}\limits}

\newcommand{\sync}{Synthetic\xspace}
\newcommand{\patternA}{PatternA\xspace}
\newcommand{\patternB}{PatternB\xspace}
\newcommand{\mocap}{MotionCapture\xspace}
\newcommand{\motes}{Motes\xspace}

\newcommand{\dcmf}{DCMF\xspace}
\newcommand{\dynammo}{DynaMMo\xspace}
\newcommand{\cdrec}{CDRec\xspace}
\newcommand{\softimpute}{SoftImpute\xspace}
\newcommand{\trmf}{TRMF\xspace}
\newcommand{\brits}{BRITS\xspace}
\newcommand{\saits}{SAITS\xspace}
\newcommand{\tider}{TIDER\xspace}

\begin{document}

\title{
Mining of Switching Sparse Networks for Missing Value Imputation in Multivariate Time Series
}


\author{Kohei Obata}
\affiliation{%
  \institution{SANKEN, Osaka University}
  \city{Osaka}
  \country{Japan}
}
\email{obata88@sanken.osaka-u.ac.jp}

\author{Koki Kawabata}
\affiliation{%
  \institution{SANKEN, Osaka University}
  \city{Osaka}
  \country{Japan}
}
\email{koki@sanken.osaka-u.ac.jp}

\author{Yasuko Matsubara}
\affiliation{%
  \institution{SANKEN, Osaka University}
  \city{Osaka}
  \country{Japan}
}
\email{yasuko@sanken.osaka-u.ac.jp}

\author{Yasushi Sakurai}
\affiliation{%
  \institution{SANKEN, Osaka University}
  \city{Osaka}
  \country{Japan}
}
\email{yasushi@sanken.osaka-u.ac.jp}


\begin{abstract}
    Multivariate time series data suffer from the problem of missing values, which hinders the application of many analytical methods.
To achieve the accurate imputation of these missing values, exploiting inter-correlation by employing the relationships between sequences (i.e., a network) is as important as the use of temporal dependency, since a sequence normally correlates with other sequences.
Moreover, exploiting an adequate network depending on time is also necessary since the network varies over time.
However, in real-world scenarios, we normally know neither the network structure nor when the network changes beforehand.
Here, we propose a missing value imputation method for multivariate time series, namely \method,
that is designed to exploit temporal dependency with a state-space model and inter-correlation by switching sparse networks.
The network encodes conditional independence between features, which helps us understand the important relationships for imputation visually.
Our algorithm, which scales linearly with reference to the length of the data,
alternatively infers networks and fills in missing values using the networks while discovering the switching of the networks.
Extensive experiments demonstrate that \method outperforms the state-of-the-art algorithms for multivariate time series imputation and provides interpretable results.

\end{abstract}






\maketitle

\section{Introduction}
    \label{010intro}
    \begin{figure}[t]
    \centering
    \begin{minipage}{1\columnwidth}
    \centering
    \includegraphics[width=1\linewidth]{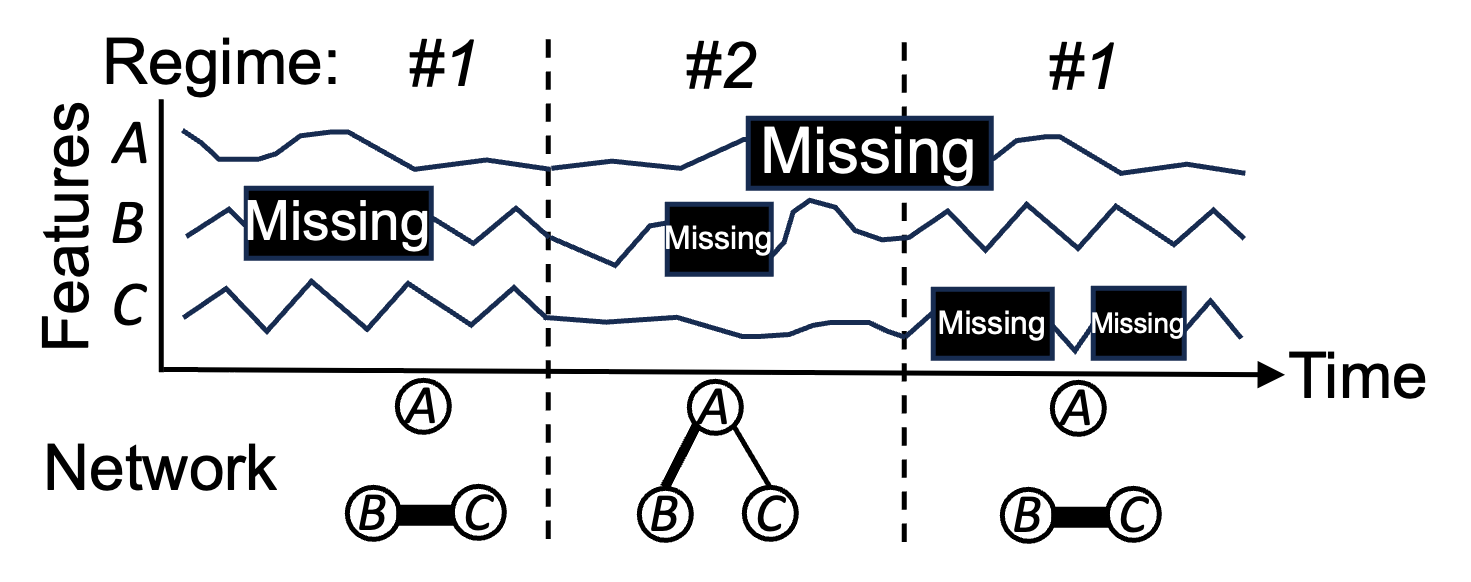} \\
    \end{minipage}
    \vspace{-1.em}
    \caption{
        An illustrative example of a multivariate time series including missing blocks, where each time point of the data is allocated to two regimes, each with a distinct network, where the edges show relationships between features.
        }
    \label{fig:concept}
    \vspace{-1.em}
\end{figure}

With the development of the Internet of Things (IoT), multivariate time series are generated in many real-world applications, such as motion capture~\cite{qin2020imaging}, and health monitoring~\cite{chambon2018deep}.
However, there are inevitably many values missing from these data, and this has many possible causes
(e.g., sensor malfunction).
As most algorithms assume an intact input when building models, missing value imputation is indispensable for real-world applications~\cite{Shadbahr_2023,berrevoets2023impute}.

In time series data, missing values often occur consecutively, leading to a missing block in a sequence, and can happen simultaneously to multiple sequences~\cite{liu2020missing}.
To effectively reconstruct missing values from such partially observed data, we must exploit both temporal dependency, by taking account of past and future values in the sequence, and inter-correlation, by using the relationship between different sequences (i.e., a network)~\cite{grin,stmvl}.
Here, a network does not necessarily mean the spatial proximity of sensors but rather underlying connectivity (e.g., Pearson correlation or partial correlation).
Moreover, as time series data are normally non-stationary, so is the network; an adequate network must be exploited depending on time~\cite{ltgl,Tomasi21icpr}.
We collectively refer to a group of time points with the same network as a ``regime''.
\fig{\ref{fig:concept}} shows an illustrative example where missing blocks randomly exist in a multivariate time series consisting of three features (i.e., \textit{A}, \textit{B}, and \textit{C}).
Each time point belongs to either of two regimes with different networks (i.e., $\#1$ and $\#2$), where the thickness of the edge indicates the strength of the interplay between features.
It is appropriate to use the values of feature \textit{C} to impute the block missing from feature \textit{B} in regime $\#1$ since the network has an edge between \textit{B} and \textit{C}.
On the other hand, in regime $\#2$, it is preferable to use feature \textit{A}, as the network suggests.
Nevertheless, in real-world scenarios, we often have no information about the data; that is, we do not know the structure of the network, let alone when the network changes.
Thus, given a partially observed multivariate time series, how can we infer networks and allocate each time point to the correct regime?
How can we achieve an accurate imputation exploiting both temporal dependency and inter-correlation?

There have been many studies on time series missing value imputation with machine learning and deep learning models~\cite{missingvldb}.
Many of them employ time-variant latent/hidden states, as used in a State-Space Model (SSM) or a Recurrent Neural Network (RNN),
to learn the dynamic patterns of time series~\cite{dynammo,btf,brits,mrnn}.
However, they do not make full use of inter-correlation, while the sequences of a multivariate time series usually interact.
Although they implicitly capture inter-correlation in latent space, they potentially capture spurious correlations under the presence of missing values, leading to inaccurate imputation. 
To tackle this problem, several studies explicitly utilize the network~\cite{dcmf,facets,netdyna,net3,grin,pogevon}.
However, they require a predefined network, even though we rarely know it in advance.

In this work, we propose \method
\footnote{Our source code and datasets are publicly available:\\
\url{https://github.com/KoheiObata/MissNet}.}
, \textit{\textbf{Mi}ning of \textbf{s}witching \textbf{s}parse \textbf{Net}works} for multivariate time series imputation, which repeatedly infers sparse networks from imputed data and fills in missing values using the networks while allocating each time point to regimes.
Specifically, our model has three components:
(1) a regime-switching model based on a discrete Markov process for detecting the change point of the network.
(2) An imputation model based on an SSM for exploiting temporal dependency and inter-correlation by latent space and the networks.
(3) A network inference model for inferring sparse networks via graphical lasso, where each network encodes pairwise conditional independencies among features, and the lasso penalty helps avoid capturing spurious correlations.
Our proposed algorithm maximizes the joint probability distribution over the above components.

Our method has the following desired properties:
\begin{itemize}
    \item \textit{Effective}: \method, which exploits both temporal dependency and inter-correlation by inferring switching sparse networks, outperforms the state-of-the-art algorithms for missing value imputation for multivariate time series.
    \item \textit{Scalable}: Our proposed algorithm scales linear with regard to the length of the time series and is thus applicable to a long-range time series.
    \item \textit{Interpretable}: \method discovers regimes with distinct sparse networks, which help us interpret data, e.g., what is the key relationship in the data, and why does a particular regime distinguish itself from others?
\end{itemize}
\section{Related work}
    \label{020related}
    We review previous studies closely related to our work.
\tabl{\ref{table:req}} shows the relative advantages of \method.
Current approaches fall short with respect to at least one of these desired characteristics.

\myparaitemize{Time series missing value imputation}
Missing value imputation for time series is a very rich topic~\cite{missingvldb}.
We roughly classify missing value imputation methods as Matrix Factorization (MF)-based, SSM-based, and Deep Learning (DL)-based approaches.

MF-based methods, such as SoftImpute~\cite{softimpute} based on Singular Value Decomposition (SVD), recover missing values from low-dimensional embedding matrices of partially observed data~\cite{spirit,cdrec,grouse,stmvl}.
For example, SoRec~\cite{sorec}, proposed as a recommendation system, constrains MF with a predefined network to exploit inter-correlation.
Since MF is limited in capturing temporal dependency, TRMF~\cite{trmf} uses an Auto-Regressive (AR) model and imposes temporal smoothness on MF.

SSMs, such as Linear Dynamical Systems (LDS)~\cite{Ghahramani1998}, use latent space to capture temporal dependency, where the data point depends on all past data points~\cite{mlds,water,btf,netdyna}.
To fit more complex time series, Switching LDS (SLDS)~\cite{pavlovic1999dynamic,fox2008nonparametric} switches multiple LDS models.
SSM-based methods, such as \dynammo~\cite{dynammo}, focus on capturing the dynamic patterns in time series rather than inter-correlation implicitly captured through the latent space.
To use the underlying connectivity in multivariate time series, \dcmf~\cite{dcmf}, and its tensor extension Facets~\cite{facets} use SSM constrained with a predefined network, which is effective, especially when the missing rate is high.
However, they assume that the network is accurately known and fixed, while it is usually unknown and may change over time in real-world scenarios.

Recently, extensive research has focused on DL-based methods, employing techniques including graph neural networks~\cite{net3,grin,stdgae,damr}, self-attention~\cite{mtand,saits}, and, most recently, diffusion models~\cite{sssd,csdi,midm}, to harness their high model capacity~\cite{gain,deepmvi}.
For example, \brits~\cite{brits} and M-RNN~\cite{mrnn} impute missing values according to hidden states from bidirectional RNN.
To utilize dynamic inter-correlation in time series, POGEVON~\cite{pogevon} requires a sequence of networks and imputes missing values in time series and missing edges in the networks, assuming that the network varies over time.
Although DL-based methods can handle complex data, the imputation quality depends heavily on the size and the selection of the training dataset.

\begin{table}[t]
    \caption{
    Capabilities of \method, Matrix Factorization (MF), State-Space Model (SSM), Deep Learning (DL), and Graphical Lasso (GL)-based approaches.
    }
    \vspace{-1em}
    \label{table:req}
    \centering
    \resizebox{1.0\linewidth}{!}{
        \begin{tabular}{c|ccc|ccc|cc|cc|c}
        \toprule
            &  \multicolumn{3}{c|}{MF} & \multicolumn{3}{c|}{SSM} & \multicolumn{2}{c|}{DL} & \multicolumn{2}{c|}{GL} & \\
            & \rotatebox{90}{\softimpute~\cite{softimpute}} & \rotatebox{90}{SoRec~\cite{sorec}} & \rotatebox{90}{\trmf~\cite{trmf}} & \rotatebox{90}{SLDS~\cite{pavlovic1999dynamic}} & \rotatebox{90}{\dynammo~\cite{dynammo}} & \rotatebox{90}{\dcmf~\cite{dcmf}} & \rotatebox{90}{\brits~\cite{brits}} & \rotatebox{90}{POGEVON~\cite{pogevon}} & \rotatebox{90}{TICC~\cite{ticc}} & \rotatebox{90}{MMGL~\cite{mmgl}} & \rotatebox{90}{\method} \\
            \midrule
            \rowcolor[gray]{0.8}
            Temporal dependency & - & - & \checkmark & \checkmark & \checkmark & \checkmark & \checkmark & \checkmark & - & - & \checkmark\\
            Inter-correlation & - & \checkmark & - & - & - & \checkmark & - & \checkmark & - & \checkmark & \checkmark\\
            \rowcolor[gray]{0.8}
            Time-varying inter-correlation & - & - & - & - & - & - & - & \checkmark & - & \checkmark & \checkmark\\
            Missing value imputation & \checkmark & \checkmark & \checkmark & - & \checkmark & \checkmark & \checkmark & \checkmark & - & \checkmark & \checkmark \\
            \rowcolor[gray]{0.8}
            Segmentation & - & - & - & \checkmark & - & - & - & - & \checkmark & - & \checkmark \\
            Sparse network inference & - & - & - & - & - & - & - & - & \checkmark & \checkmark & \checkmark \\
            \bottomrule
        \end{tabular}
    }
    \vspace{-1em}
\end{table}

\myparaitemize{Sparse network inference}
From another perspective, our method infers sparse networks from time series containing missing values and discovers regimes (i.e., clusters) based on networks.
Inferring a sparse inverse covariance matrix (i.e., network) from data helps us to understand feature dependency in a statistical way.
Graphical lasso~\cite{graphicallasso}, which maximizes the Gaussian log-likelihood imposing an \lonenorm penalty,
is one of the most commonly used techniques for estimating a sparse network from static data.
St\"{a}dler and B\"{u}hlmann~\cite{missgl} have tackled inferring a sparse network from partially observed data according to conditional probability.
However, the network varies over time~\cite{NAMAKI20113835,monti2014estimating,cpdgl}; thus, TVGL~\cite{tvgl} infers time-varying networks by considering the time similarity with a network belonging to neighboring segments.
To infer time-varying networks in the presence of missing values, MMGL~\cite{mmgl}, which employs TVGL, uses the expectation-maximization (EM) algorithm to repeat the inference of time-varying networks and missing value imputation based on conditional probability under the condition that each segment has the same observed features.
Discovering clusters based on networks~\cite{dmm}, such as TICC~\cite{ticc} and TAGM~\cite{tagm}, provides interpretable results that other traditional clustering methods cannot find.
However, they cannot handle missing values.

As a consequence, none of the previous studies have addressed missing value imputation for multivariate time series by employing sparse network inference and segmentation based on the network.

\section{Preliminaries}
    \label{030preliminary}
    \subsection{Problem definition}
In this paper, we focus on the task of multivariate time series missing value imputation.
We use a multivariate time series with $\ndim$ features and $\timestep$ timesteps $\mts = \{ \mtsvector_1,\mtsvector_2,\dots,\mtsvector_\timestep \} \in \realnumber^{\ndim \times \timestep}$.
To represent the missing values in $\mts$, we introduce the indicator matrix $\indicatormatrix \in \realnumber^{\ndim \times \timestep}$, 
where $\indicatormatrix_{i,t}$ indicates the availability of feature $i$ at timestep $t$:
$\indicatormatrix_{i,t}$ being $0$ or $1$ indicates whether $\mts_{i,t}$ is missing or observed.
Thus, the observed entries can be described as $\givenmts = \indicatormatrix \hadamard \mts$, where $\hadamard$ is a Hadamard product.
Our problem is formally written as follows:
\begin{problem}[Multivariate Time Series Missing Value Imputation] \label{prob:}
\textbf{Given}: a partially observed multivariate time series $\givenmts$;
\textbf{Recover}: its missing parts indicated by $\indicatormatrix$.
\end{problem}

\subsection{Graphical lasso}
We use graphical lasso~\cite{graphicallasso} to infer the network for each regime.
Given $\mts$, graphical lasso estimates the sparse Gaussian inverse covariance matrix (i.e., network) $\invcovmatrix \in \realnumber^{\ndim \times \ndim}$, also known as the precision matrix.
The network encodes pairwise conditional independencies among $\ndim$ features,
e.g., if $\invcovmatrix_{i,j} = 0$, then features $i$ and $j$ are conditionally independent given the values of all the other features.
The optimization problem is expressed as follows:
\begin{align}
\textrm{minimize}_{\invcovmatrix \in S^{p}_{++}}
    &\sparseparam||\invcovmatrix||_{od,1} - \sum_{t=1}^{\timestep} \loglike(\mtsvector_{t,},\invcovmatrix), \label{eq:gl} \\
\loglike(\mtsvector_{t},\invcovmatrix) = &-\frac{1}{2} (\mtsvector_{t} - \meanvector)' \invcovmatrix (\mtsvector_{t} - \meanvector) \nonumber \\
    &+ \frac{1}{2} \log \textrm{det}(\invcovmatrix) -\frac{\ndim}{2} \log(2\pi) , \label{eq:ll}
\end{align}
where $\invcovmatrix$ must be symmetric positive definite ($S^{p}_{++}$).
$\loglike(\mtsvector_{t}, \invcovmatrix)$ is the log-likelihood and $\meanvector \in \realnumber^{\ndim}$ is the empirical mean of $\mts$.
$\sparseparam \geq 0$ is a hyperparameter for determining
the sparsity level of the network,
and $\|\cdot\|_{od,1}$ indicates the off-diagonal \lonenorm.
This is a convex optimization problem that can be solved via the alternating direction method of multipliers (ADMM)~\cite{admm}.

\section{Proposed \method}
    \label{040model}
    In this section, we present our proposed \method for missing value imputation.
We give the formal formulation of the model and then provide the detailed algorithm to learn the model.

\subsection{Optimization model}
\method infers sparse networks and fills in missing values using the networks while discovering regimes.
We first introduce three interacting components of our model:
a regime-switching model, an imputation model, and a network inference model.
Then, we define the optimization formulation.

\myparaitemize{Regime-switching model}
\method describes the change of networks by regime-switching.
Let $\nstate$ be the number of regimes,
$\switchmatrix = \{ \switchvector_{1}, \dots \switchvector_\timestep \} \in \realnumber^{\nstate \times \timestep}$ be regime assignments,
and $\switchvector_{t}$ be a one-hot vector
\footnote{We use it interchangeably with the index of the regime
(i.e., \thN{\switchvector_{t}}-regime).}
that indicates $\givenmtsvector_{t} \in \realnumber^{\ndim}$ belongs to \thN{\switchvector_{t}}-regime.
We assume regime-switching to be a discrete first-order Markov process:
\begin{align}
    \probability{\switchvector_{1}} = \mcini, \quad
    \conditionprobability{\switchvector_{t+1}}{\switchvector_{t}} = \mctra_{\switchvector_{t+1},\switchvector_{t}},
\end{align}
where $\mctra \in \realnumber^{\nstate \times \nstate}$ is the Markov transition matrix and $\mcini \in \realnumber^{\nstate}$ is the initial state distribution.

\myparaitemize{Imputation model}
\method imputes missing values exploiting temporal dependency and inter-correlation indicated by the networks.
We assume that the temporal dependency is consistent throughout all regimes and captured in the latent space of an SSM, which allows us to consider long-term dependency.
%
We define the latent states $\latentfactor = \{ \latentfactorvector_1,\latentfactorvector_2,\dots,\latentfactorvector_\timestep \} \in \realnumber^{\latentdim \times \timestep}$
corresponding to $\givenmts$, where $\latentdim$ is the number of latent dimensions,
and $\latentfactorvector_{t+1} \in \realnumber^{\latentdim}$
is linear to $\latentfactorvector_{t}$
through the transition matrix $\transitionmatrix \in \realnumber^{\latentdim \times \latentdim}$
with the covariance $\SgmZ$,
shown in \eq{\ref{eq:latent_state}}.
%
As defined in \eq{\ref{eq:initial_state}},
the first timestep of latent state $\latentfactorvector_{1}$
is defined by the initial state $\latentfactorvector_{0}$
and the covariance $\sgmZero$.
\begin{align}
    \latentfactorvector_{1} \sim & \gauss{\latentfactorvector_{0}}{\sgmZero}, \label{eq:initial_state} \\
    \latentfactorvector_{t+1} | \latentfactorvector_{t} \sim & \gauss{\transitionmatrix \latentfactorvector_{t}}{\SgmZ}. \label{eq:latent_state}
\end{align}

We define the observation $\givenmtsvector_{t}$
assigned to \thN{\switchvector_{t}}-regime
as being linear to the latent state $\latentfactorvector_{t}$
through the observation matrix of \thN{\switchvector_{t}}-regime $\objectmatrixN{\switchvector_{t}} \in \realnumber^{\ndim \times \latentdim}$
with the covariance $\SgmXN{\switchvector_{t}}$:
\begin{align}
    \givenmtsvector_{t} | \latentfactorvector_{t}, \objectmatrixN{\switchvector_{t}}, \switchvector_{t} &\sim \gauss{\objectmatrixN{\switchvector_{t}} \latentfactorvector_{t}}{\SgmXN{\switchvector_{t}}}. \label{eq:observation}
\end{align}

\method captures inter-correlation by adding a constraint on $\objectmatrixN{k}$.
Let it be assumed that the contextual matrix of the \thN{k}-regime $\contextualmatrixN{k} \in \realnumber^{\ndim \times \ndim}$ encodes the inter-correlation of $\mts$ belonging to the \thN{k}-regime ($1 \leq k \leq \nstate$).
We define the contextual latent factor of the \thN{k}-regime $\objectfactorN{k} \in \realnumber^{\latentdim \times \ndim}$,
and the $j$-th column ($1 \leq j \leq \ndim$) of the contextual matrix $\contextualvectorN{k}_{j}$
as linear to $\objectfactorvectorN{k}_{j}$
through the observation matrix $\objectmatrixN{k}$
with the covariance $\SgmSN{k}$,
formulated in \eq{\ref{eq:contextual_matrix}}.
%
In \eq{\ref{eq:contextual_latent_matrix}},
we place zero-mean spherical Gaussian priors on $\objectfactorvectorN{k}_{j}$
under the assumption that $-1 \leq \contextualmatrixN{k}_{i,j} \leq 1$.
\begin{align}
    \contextualvectorN{k}_{j} | \objectfactorvectorN{k}_{j}, \objectmatrixN{k} &\sim \gauss{\objectmatrixN{k} \objectfactorvectorN{k}_{j}}{\SgmSN{k}}, \label{eq:contextual_matrix} \\
    \objectfactorvectorN{k}_{j} &\sim \gauss{0}{\SgmVN{k}}. \label{eq:contextual_latent_matrix}
\end{align}

To avoid our model overfitting the observed entries since many entries are missing,
we simplify the covariances by assuming that all noises are independent and identically distributed (i.i.d.).
Thus, the parameters $\SgmZ, \SgmXN{k}, \SgmSN{k}, \SgmVN{k}$
are reduced to $\sgmZ^{2}, \sgmXN{k}^{2}, \sgmSN{k}^{2}, \linebreak \sgmVN{k}^{2}$, respectively
(i.e., $\SgmZ =\sgmZ^{2}\idmat$, $\SgmXN{k} =\sgmXN{k}^{2}\idmat$, $\SgmSN{k} =\sgmSN{k}^{2}\idmat$, and $\SgmVN{k} =\sgmVN{k}^{2}\idmat$).

With the above imputation model,
the imputed time series $\impmtsvector_{t}$ at timestep $t$ is recovered as follows:
\begin{align}
    \impmtsvector_{t} = \indicatormatrix_{:,t} \hadamard \givenmtsvector_{t} + (1-\indicatormatrix_{:,t}) \hadamard \objectmatrixN{\switchvector_{t}} \latentfactorvector_{t} . \label{eq:imputation}
\end{align}

\myparaitemize{Network inference model}
\method infers the network for each regime to exploit inter-correlation.
We define a Gaussian distribution and \lonenorm for the imputed data belonging to each regime to allow us to estimate networks accurately:
\begin{align}
    \impmtsvector_{t} | \switchvector_{t} \sim \gauss{\meanvectorN{\switchvector_{t}}}{\invcovmatrix^{(\switchvector_{t})-1}}, \:\:\:
    s.t., ||\invcovmatrixN{\switchvector_{t}}||_{od,1} \leq \frac{\constant}{\sparseparam}, \label{eq:network_inference}
\end{align}
where $\constant$ is any constant value for convenience,
and $\sparseparam$ is a hyperparameter that controls the sparsity of the network (i.e., inverse covariance matrix) $\invcovmatrixN{\switchvector_{t}}$ with \lonenorm, 
which helps avoid capturing spurious correlations.

\myparaitemize{Optimazation formulation}
Our goal is to estimate the model parameters $\parameter = \{ \transitionmatrix, \latentfactorvector_{0}, \sgmZero, \sgmZ, \sgmXset, \sgmSset, \sgmVset,  \meanvectorset, \invcovmatrixset, \mcini, \mctra \}$
and find the latent factors $\latentfactor, \objectfactorset, \objectmatrixset, \contextualmatrixset, \switchmatrix$,
where the letters with a dot indicate a set of $\nstate$ vectors/matrices/scalers (e.g., $\contextualmatrixset = \setK{\contextualmatrixN{k}}$),
that maximizes the following joint probability distribution:
\begin{align}
    & \argmax \probability{\givenmts, \latentfactor, \objectfactorset, \objectmatrixset, \contextualmatrixset, \switchmatrix} = 
    \underbrace{ \probability{\latentfactorvector_{1}}
    \prod_{t=2}^{\timestep} \conditionprobability{\latentfactorvector_{t}}{\latentfactorvector_{t-1}} }_{\text{temporal dependency}} \nonumber \\
    & \underbrace{ \prod_{t=1}^{\timestep} \conditionprobability{\givenmtsvector_{t}}{\latentfactorvector_{t-1},\objectmatrixN{\switchvector_{t}},\switchvector_{t}} }_{\text{time series}}
    \underbrace{ \prod_{k=1}^{\nstate} \Bigg\lbrace \prod_{j=1}^{\ndim} \probability{\objectfactorvectorN{k}_j}  \prod_{j=1}^{\ndim} \conditionprobability{\contextualvectorN{k}_j}{\objectmatrixN{k},\objectfactorvectorN{k}_j} \Bigg\rbrace}_{\text{network constraint}} \nonumber \\
    & \underbrace{ \probability{\switchvector_{1}}
    \prod_{t=2}^{\timestep} \conditionprobability{\switchvector_{t}}{\switchvector_{t-1}} }_{\text{regime-switching}}
    \underbrace{ \prod_{t=1}^{\timestep} \conditionprobability{\impmtsvector_{t}}{\meanvectorN{\switchvector_{t}},\invcovmatrixN{\switchvector_{t}},\switchvector_{t}} }_{\text{network inference}}, \nonumber \quad
    \underbrace{ ( s.t. ||\invcovmatrixN{k}||_{od,1} \leq \frac{\constant}{\sparseparam} )}_{\text{network sparsity}} . \label{eq:joint} \\
\end{align}

\begin{figure}[t]
    \centering
    \begin{minipage}{1\columnwidth}
    \centering
    \includegraphics[width=.9\linewidth]{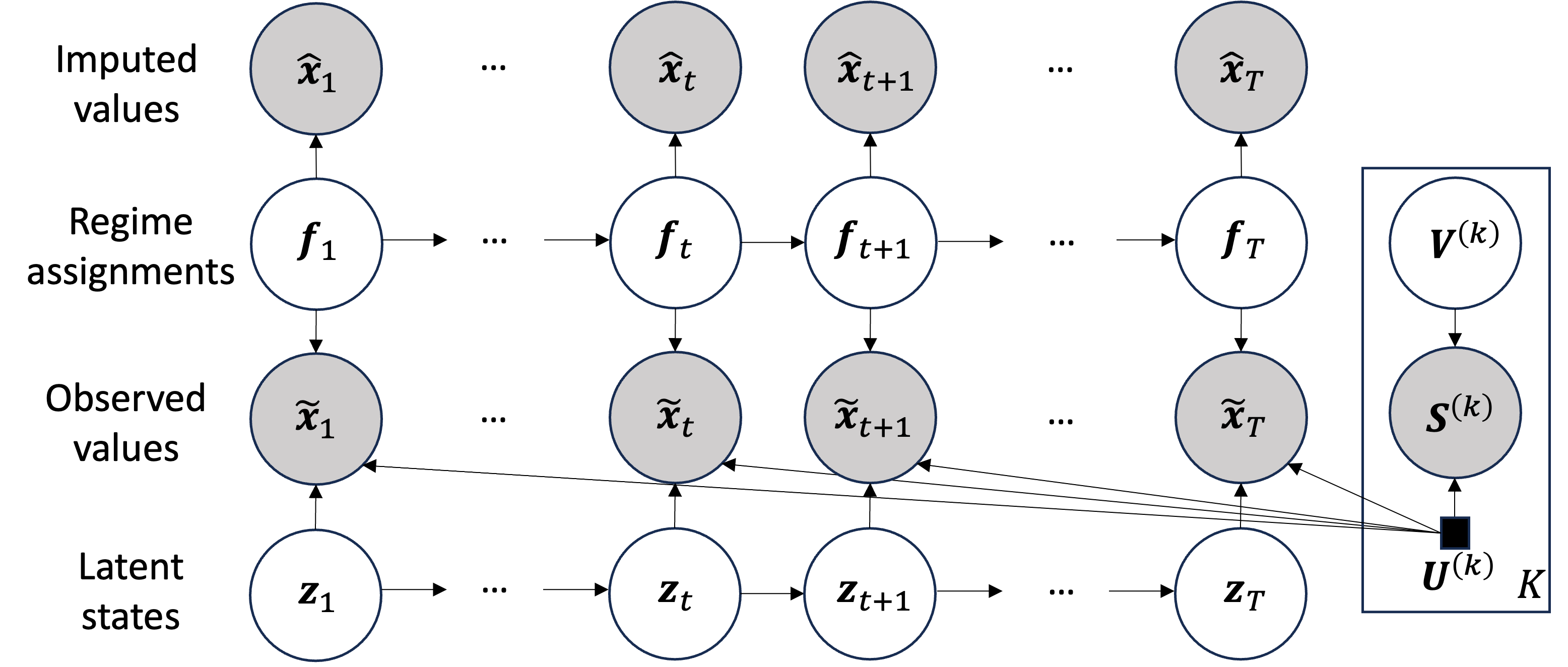} \\
    \end{minipage}
    \vspace{-1em}
    \caption{
        Graphical model of \method at each iteration.
        }
    \label{fig:graphical}
    \vspace{-1em}
\end{figure}

\subsection{Algorithm}
It is difficult to find the global optimal solution of \eq{\ref{eq:joint}}, for the following reasons:
\romanone As a constraint for $\objectmatrixset$, $\contextualmatrixset$ has to be fixed and encode inter-correlation;
\romantwo $\objectmatrixset$, $\latentfactor$ and $\switchmatrix$ jointly determine $\givenmts$, and $\objectmatrixset$ and $\objectfactorset$ jointly determine $\contextualmatrixset$;
\romanthree Calculating the correct $\switchmatrix$ is NP-hard.
Hence, we aim to find its local optimum instead, following the EM algorithm, where the graphical model for each iteration is shown in \fig{\ref{fig:graphical}}.
Specifically, to address the aforementioned difficulties, we employ the following steps:
\romanone we consider $\contextualmatrixset$ is observed in each iteration, and we update it at the end of the iteration using $\invcovmatrixset$;
\romantwo we regard $\objectmatrixset$ as a model parameter, 
thus $\{ \givenmts, \latentfactor, \switchmatrix \}$ are independent with $\{ \contextualmatrixset, \objectfactorset \}$.
We alternate the inference of $\{ \latentfactor, \switchmatrix \}$ and $\objectfactorset$ and the update of the model parameters;
\romanthree we employ a Viterbi approximation and infer the most likely $\switchmatrix$.

\subsubsection{E-step}
Given $\objectmatrixset$, we can infer $\{ \latentfactor, \switchmatrix \}$ and $\objectfactorset$ independently.

Let $\obsindex_{t}$ denote the indices of the observed entries of $\givenmtsvector_{t}$.
The observed-only data $\obsmtsvector_{t}$ and the corresponding observed-only observation matrix $\obsobjectmatrixN{k}_{t}$ are defined as follows:
\begin{align}
    \obsindex_{t} = \{i | \indicatormatrix_{i,t} > 0, i= 1,\dots,\ndim \} , \nonumber \\
    \obsmtsvector_{t} = \givenmtsvector_{t}(\obsindex_{t}), \quad
    \obsobjectmatrixN{k}_{t} = \objectmatrixN{k}(\obsindex_{t}, :) .
\end{align}

\myparaitemize{Inferring $\switchmatrix$ and $\latentfactor$}
$\switchmatrix$ and $\latentfactor$ are coupled, and so must be jointly determined.
We first use a Viterbi approximation to find the most likely regime assignments $\switchmatrix$ that maximize the log-likelihood \eq{\ref{eq:joint}}.
The likelihood term obtained during the calculation of $\switchmatrix$ also acts as a Kalman Filter (forward algorithm).
Then, we infer $\latentfactor$ with a Rauch-Tung-Streibel (RTS) smoother (backward algorithm).

In a Viterbi approximation, finding $\switchmatrix$ requires the partial cost $\switchcost_{t,t-1,k,l}$ when the switch is to regime $k$ at time $t$ from regime $l$ at time $t-1$.
To calculate the partial cost,
we define the following LDS state and variance terms:
\begin{align}
    \formean_{t|t-1,k,l} &= \transitionmatrix \formean_{t-1|t-1,k,l}, \nonumber \\
    \formean_{t|t,k,l} &= \formean_{t|t-1,k,l} + \kalmangain_{t,k,l} (\obsmtsvector_{t} - \obsobjectmatrixN{k}_{t} \transitionmatrix \formean_{t|t-1,k,l}), \nonumber \\
    \forcov_{t|t-1,k,l} &= \transitionmatrix \forcov_{t-1|t-1,k,l} \transitionmatrix' + \sgmZ^{2} \idmat, \nonumber \\
    \forcov_{t|t,k,l} &= (\idmat - \kalmangain_{t,k,l} \obsobjectmatrixN{k}_{t})\forcov_{t-1|t-1,k,l}, \nonumber \\
    \kalmangain_{t,k,l} &= \forcov_{t|t-1,k,l} \obsobjectmatrixNinv{k}_{t} ( \obsobjectmatrixN{k}_{t} \forcov_{t|t-1,k,l} \obsobjectmatrixNinv{k}_{t} + \sgmXN{k}^{2} \idmat)^{-1} , \label{eq:forward}
\end{align}
with the initial state:
\begin{align}
    \formean_{1|1,k} &= \latentfactorvector_{0} + \kalmangain_{1,k}(\obsmtsvector_{1} - \obsobjectmatrixN{k}_{1} \latentfactorvector_{0}) , \nonumber \\
    \forcov_{1|1,k} &= (\idmat - \kalmangain_{1,k} \obsobjectmatrixN{k}_{1})\sgmZero , \nonumber \\
    \kalmangain_{1,k} &= \sgmZero \obsobjectmatrixNinv{k}_{1} ( \obsobjectmatrixN{k}_{1} \sgmZero \obsobjectmatrixNinv{k}_{1} + \sgmXN{k}^{2} \idmat)^{-1} , \label{eq:initial}
\end{align}
where $\formean_{t|t-1,k,l}$ and $\formean_{t|t,k,l}$ are the one-step predicted LDS state and the best-filtered state estimates at $t$, respectively, given the switch is in regime $k$ at time $t$ and in regime $l$ at time $t-1$ and only the $t-1$ measurements are known.
Similar definitions are used for $\forcov_{t|t-1,k,l}$ and $\forcov_{t|t,k,l}$.

The partial cost is obtained by calculating the logarithm of \eq{\ref{eq:joint}} related to $\switchvector_{t}$:
\begin{align}
    \switchcost_{t,t-1,k,l} =& \frac{1}{2} (\obsmtsvector_{t} - \obsobjectmatrixN{k}_{t} \formean_{t|t-1,k,l})' \invsig (\obsmtsvector_{t} - \obsobjectmatrixN{k}_{t} \formean_{t|t-1,k,l}) \nonumber \\
    &- \frac{1}{2}\log \textrm{det}(\invsig) + \frac{\latentdim}{2} log(2 \pi) \nonumber \\
    &+ \frac{1}{2} (\impmtsvector_{t} - \meanvectorN{k})' \invcovmatrixN{k} (\impmtsvector_{t} - \meanvectorN{k}) \nonumber \\
    &- \frac{1}{2} \log \textrm{det} (\invcovmatrixN{k}) + \frac{\ndim}{2} \log(2\pi) - \log(\mctra_{k,l}) , \label{eq:partialcost} \\
    \invsig =& (\obsobjectmatrixN{k}_{t} \forcov_{t|t-1,k,l} \obsobjectmatrixNinv{k}_{t} + \sgmXN{k}^{2} \idmat)^{-1} . \nonumber 
\end{align}

Once all partial costs are obtained, it is well-known how to apply a Viterbi inference to a discrete Markov process to obtain the most likely regime assignments $\switchmatrix$~\cite{viterbi}.

Then, we infer $\latentfactor$.
Let the posteriors of $\latentfactorvector_{t}$ be as follows:
\begin{align}
    \conditionprobability{\latentfactorvector_{t}}{\givenmtsvector_{1}, \dots, \givenmtsvector_{t}} = \gauss{\latentfactorvector_{t}|\formean_{t}}{\forcov_{t}} , \nonumber \\
    \conditionprobability{\latentfactorvector_{t}}{\givenmtsvector_{1}, \dots, \givenmtsvector_{\timestep}} = \gauss{\latentfactorvector_{t}|\bacmean_{t}}{\baccov_{t}} .
\end{align}

We now obtain $\formean_{t}, \forcov_{t}, \forcov_{t|t-1}$ using $\switchmatrix$; thus, in practice, we have conducted a Kalman Filter.
We apply the RTS smoother to infer $\latentfactor$.
\begin{align}
    \bacmean_{t} &= \formean_{t} + \kalmanQ_{t}(\bacmean_{t+1} - \transitionmatrix \formean_{t}) , \nonumber \\
    \baccov_{t} &= \forcov_{t} + \kalmanQ_{t}(\baccov_{t+1} - \forcov_{t|t-1})\kalmanQ'_{t} , \nonumber \\
    \kalmanQ_{t} &= \forcov_{t} \transitionmatrix' (\forcov_{t|t-1})^{-1} , \label{eq:backward} \\
    \expectation{\latentfactorvector_{t}} &= \bacmean_{t} , \nonumber \\
    \expectation{\latentfactorvector_{t} \latentfactorvector'_{t-1}} &= \baccov_{t}\kalmanQ'_{t} + \bacmean_{t}\bacmean'_{t-1} , \nonumber \\
    \expectation{\latentfactorvector_{t} \latentfactorvector'_{t}} &= \baccov_{t} + \bacmean_{t}\bacmean'_{t-1} . \label{eq:latentfactor}
\end{align}

\myparaitemize{Inferring $\objectfactorset$}
We apply Bayes' theorem to \eq{\ref{eq:contextual_matrix}} and (\ref{eq:contextual_latent_matrix}) to obtain the posteriors
$ \conditionprobability{\objectfactorvectorN{k}_{j}}{\contextualvectorN{k}_{j}} = \gauss{\objectfactorvectorN{k}_{j}|\baymeanN{k}_{j}}{\baycovN{k}} $:
\begin{align}
    \baymeanN{k}_{j} &= \bayM^{(k)-1} \objectmatrix^{(k)'} \contextualvectorN{k}_{j} , \nonumber \\
    \baycovN{k} &= \sgmSN{k}^{2} \bayM^{(k)-1} , \nonumber \\
    \bayMN{k} &= \objectmatrix^{(k)'}\objectmatrixN{k} + \sgmVN{k}^{-2} \sgmSN{k}^{2} \idmat , \nonumber \\
    \expectation{\objectfactorvectorN{k}_{j}} &= \baymeanN{k}_{j} , \nonumber \\
    \expectation{\objectfactorvectorN{k}_{j}\objectfactorvector^{(k)'}_{j}} &= \baycovN{k} + \baymeanN{k}_{j}\baymean^{(k)'}_{j} . \label{eq:objectmatrix}
\end{align}

\subsubsection{M-step}
After obtaining $\{ \latentfactor, \switchmatrix \}$ and $\objectfactorset$, we update the model parameters to maximize the expectation of the log-likelihood:
\begin{align}
    \parameter^{new} &= \argmax_{\parameter} \objective{\parameter, \parameter^{old}} , \label{eq:parameter} \\
    \objective{\parameter, \parameter^{old}} &= \expectationunder{\log \probability{\givenmts, \latentfactor, \objectfactorset, \objectmatrixset, \contextualmatrixset, \switchmatrix|\parameter}}{\latentfactor,\switchmatrix,\objectfactorset|\parameter^{old}} - \sum_{k=1}^{\nstate}\sparseparam||\invcovmatrixN{k}||_{od,1} , \nonumber 
\end{align}
where we incorporate the \lonenorm constraint.

Parameters for the imputation model, $\objectmatrixset$ and $\{ \transitionmatrix, \latentfactorvector_{0}, \sgmZero, \sgmZ, \sgmXset, \sgmSset, \sgmVset \}$,
can be obtained by taking the derivative of $\objective{\parameter, \parameter^{old}}$.
For $\objectmatrixN{k}$, we update each row $\objectmatrixN{k}_{i,:}$ individually:
\begin{align}
    (\objectmatrixN{k}_{i,:})^{new} &= \UoneN{k} \Utwo^{(k)-1} , \label{eq:paramobject} \\
    \UoneN{k} = \contextparam \sgmSN{k}^{-2} \sum_{j=1}^{\ndim} \contextualmatrixN{k}_{i,j} \expectation{\objectfactorvector^{(k)'}_{j}} 
    &+ (1-\contextparam) \sgmXN{k}^{-2} \sum_{t=1, \switchvector_{t} \in k}^{\timestep} \indicatormatrix_{i,t} \givenmts_{i,t} \expectation{\latentfactorvector'_{t}}, \nonumber \\
    \UtwoN{k} = \contextparam \sgmSN{k}^{-2} \sum_{j=1}^{\ndim} \expectation{\objectfactorvector^{(k)}_{j} \objectfactorvector^{(k)'}_{j}} 
    &+ (1-\contextparam) \sgmXN{k}^{-2} \sum_{t=1, \switchvector_{t} \in k}^{\timestep} \indicatormatrix_{i,t} \expectation{\latentfactorvector_{t} \latentfactorvector'_{t}}, \nonumber 
\end{align}
where $0 \leq \contextparam \leq 1$ is a hyperparameter employed as a trade-off for the contributions of inter-correlation and temporal dependency.
The details of updating $\{ \transitionmatrix, \latentfactorvector_{0}, \sgmZero, \sgmZ, \sgmXset, \sgmSset, \sgmVset \}$ are presented in \apdx{\ref{apd:algparam}}.

For the network inference model, we calculate $\impmts$ with \eq{\ref{eq:updateimpute}} and then update $\meanvectorN{k}$ by calculating the empirical mean of $\impmts$ belonging to the \thN{k}-regime
and $\invcovmatrixN{k}$ by solving the graphical lasso problem shown in \eq{\ref{eq:network}} via ADMM.
\begin{align}
    \impmtsvector_{t} = \indicatormatrix_{:,t} \hadamard \givenmtsvector_{t} + (1-\indicatormatrix_{:,t}) \hadamard (\objectmatrixN{\switchvector_{t}})^{new} \bacmean_{t}, \label{eq:updateimpute} \\
    \textrm{minimize}_{\invcovmatrixN{k} \in S^{p}_{++}}
    \sparseparam||\invcovmatrixN{k}||_{od,1} - \sum_{t=1, \switchvector_{t} \in k}^{\timestep} \loglike(\impmtsvector_{t},\invcovmatrixN{k}). \label{eq:network}
\end{align}

For the regime-switching model, the initial state distribution and the Markov transition matrix are updated as follows:
\begin{align}
    \mcini = \switchvector_{1} , \quad
    \mctra = \bigg( \sum_{t=2}^{\timestep} \switchvector_{t}\switchvector_{t-1}' \bigg) \textrm{diag} \bigg( \sum_{t=2}^{\timestep} \switchvector_{t} \bigg)^{-1} . \label{eq:paramswitch}
\end{align}

\subsubsection{Update $\contextualmatrixset$}
We update $\contextualmatrixN{k}$ at the end of each iteration.
As shown in \eq{\ref{eq:paramcontext}},
we define the off-diagonal elements of $\contextualmatrixN{k}$ as partial correlations calculated from the network $\invcovmatrixN{k}$ to encode the inter-correlation.
\begin{align}
    \contextualmatrixN{k}_{i,j} &=
    \begin{cases}
        1 & (i = j) \\
        -(\frac{\invcovmatrixN{k}_{i,j} }{ \sqrt{\invcovmatrixN{k}_{i,i}} \sqrt{\invcovmatrixN{k}_{j,j}} }) &(i \neq j) 
    \end{cases}. \label{eq:paramcontext}
\end{align}

\begin{algorithm}[t]
    \caption{\textsc{\method($\givenmts, \indicatormatrix, \latentdim, \nstate, \contextparam, \sparseparam$)}}
    \label{alg:all}
    \begin{algorithmic}[1]
        \STATE {\bf Input:}
        (a) partially observed multivariate time series $\givenmts$, \\
        \hspace{2.75em} (b) indicator matrix $\indicatormatrix$, \\
        \hspace{2.75em} (c) and hyperparameters $\latentdim, \nstate, \contextparam, \sparseparam$
        \STATE {\bf Output:} latent factors $\latentfactor, \objectfactorset, \objectmatrixset, \contextualmatrixset, \switchmatrix$
        model parameters $\parameter$,
        and $\impmts$
        \STATE Initialize $\impmts$ with linear interpolation, $\latentfactor, \objectfactorset, \objectmatrixset, \contextualmatrixset, \switchmatrix$, and $\parameter$; \label{step:init}
        \REPEAT
        \FOR{$t=1:\timestep$} \label{step:viterbi_start}
            \FOR{$k=1:\nstate$}
                \FOR{$l=1:\nstate$}
                    \STATE Calculate partial cost $\switchcost_{t,t-1,k,l}$ using \eq{\ref{eq:partialcost}} \label{step:viterbi} 
                \ENDFOR
            \ENDFOR
        \ENDFOR
        \STATE Infer $\switchmatrix$ and obtain $\formean_{t}, \forcov_{t}$ based on \eq{\ref{eq:forward}}, (\ref{eq:initial}) \label{step:viterbi_end}
        \FOR{$t=\timestep:1$} \label{step:z_start}
            \STATE Infer $\bacmean_{t}, \baccov_{t}, \expectation{\latentfactorvector_{t} \latentfactorvector'_{t-1}}, \expectation{\latentfactorvector_{t} \latentfactorvector'_{t}}$ by \eq{\ref{eq:backward}}, (\ref{eq:latentfactor}) \label{step:backward} 
        \ENDFOR \label{step:z_end}
        \FOR{$j=1:\ndim$} \label{step:v_start}
            \STATE Infer $\expectation{\objectfactorvectorN{k}_{j}}, \expectation{\objectfactorvectorN{k}_{j}\objectfactorvector^{(k)'}_{j}}$ using \eq{\ref{eq:objectmatrix}} \label{step:bayes}
        \ENDFOR \label{step:v_end}
        \STATE Set $\latentfactor = \{ \bacmean_{1}, \dots, \bacmean_{\timestep} \}$, $\setK{\objectfactorN{k} = \{ \baymeanN{k}_{1}, \dots, \baymeanN{k}_{\ndim} \}}$
        \STATE Update $\objectmatrixset$, $\parameter$ and $\impmts$ by \eq{\ref{eq:paramobject}} - (\ref{eq:paramswitch}) \label{step:param}
        \STATE Update $\contextualmatrixset$ based on \eq{\ref{eq:paramcontext}} \label{step:context}
        \UNTIL{convergence;}
        \RETURN $\{ \latentfactor, \objectfactorset, \objectmatrixset, \contextualmatrixset, \switchmatrix, \parameter, \impmts \}$;
    \end{algorithmic}
    \normalsize
\end{algorithm}

\subsubsection{Overall algorithm}
We have the overall algorithm shown as \alg{\ref{alg:all}} to obtain a local optimal solution of \eq{\ref{eq:joint}}.
Given a partially observed multivariate time series $\givenmts$,
an indicator matrix $\indicatormatrix$,
the dimension of latent state $\latentdim$,
the number of regimes $\nstate$,
the network parameter $\contextparam$,
and the sparse parameter $\sparseparam$,
our algorithm aims to find the latent factors $\latentfactor, \objectfactorset, \objectmatrixset, \contextualmatrixset, \switchmatrix$,
other model parameters in $\parameter$,
and imputed time series $\impmts$.

The \method algorithm starts by initializing $\impmts$ with a linear interpolation, and by randomly initializing $\latentfactor, \objectfactorset, \objectmatrixset, \contextualmatrixset, \switchmatrix$, and $\parameter$ (\step{\ref{step:init}}).
Then, it alternately updates the latent factors and parameters until they converge.
In each iteration, we consider $\contextualmatrixset$ to be given and $\objectmatrixset$ to be a model parameter.
In an iteration, we first conduct a Viterbi approximation to calculate the most likely regime assignments $\switchmatrix$ (\step{\ref{step:viterbi_start}-\ref{step:viterbi_end}}).
Then, we infer the expectations of $\latentfactor$ and $\objectfactorset$ (\step{\ref{step:z_start}-\ref{step:z_end}} and \step{\ref{step:v_start}-\ref{step:v_end}}),
and we update $\objectmatrixset$ and model parameters $\parameter$ (\step{\ref{step:param}}),
and at the end of the iteration, we update $\contextualmatrixset$ (\step{\ref{step:context}}).

\subsection{Complexity analysis}
\begin{lemma} \label{lemma:time}
    The time complexity of \method is \\
    $O( \#iter \cdot ( \nstate^{2}\sum_{t=1}^\timestep(\latentdim^{3} + \latentdim^{2}\ndim_{t} + \latentdim\ndim_{t}^{2} + \ndim_{t}^{3}) + \nstate\latentdim^{2}\ndim^{2} + \nstate\timestep\latentdim^{2}\ndim + \nstate\ndim^{3} ) )$.
\end{lemma}
\begin{proof}
    See \apdx{\ref{apd:proof}}.
\end{proof}
$\ndim_{t}$ represents the number of observed features of $\obsmtsvector_{t}$. 
Note that $\nstate^{2}\sum_{t=1}^\timestep(\latentdim^{3} + \latentdim^{2}\ndim_{t} + \latentdim\ndim_{t}^{2} + \ndim_{t}^{3})$ is upper bounded by $\nstate^{2} \timestep \ndim^{3}$.
In practice, the length of the time series ($\timestep$) is often orders of magnitude greater than the number of features ($\ndim$).
Hence, the actual running time of \method is dominated by the term related to $\timestep$, which is linear in $\timestep$.

\begin{lemma} \label{lemma:space}
    The space complexity of \method is \\
    $O( \timestep\ndim + \nstate^{2}\timestep\latentdim^{2} + \nstate\latentdim^{2}\ndim + \nstate\ndim^{2} )$.
\end{lemma}
\begin{proof}
    See \apdx{\ref{apd:proof}}.
\end{proof}
\section{Experiments}
    \label{050experiments}
    
\begin{figure*}[t]
    \centering
    \begin{tabular}{l}
    \includegraphics[width=1.00\linewidth]{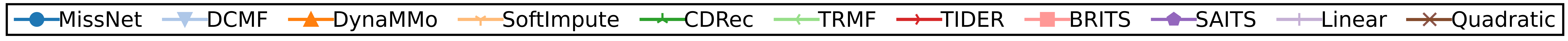} \\
    \end{tabular}

    \begin{minipage}{1.00\linewidth}
    \centering
    \begin{tabular}{cccccc}
    \hspace{-1.000em}
    \begin{minipage}{0.15\linewidth}
    \centering
    \includegraphics[width=1\linewidth]{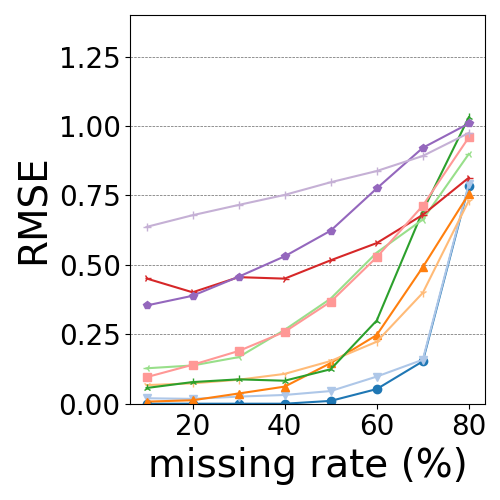} \\
    \vspace{-.5em}
    (a) \patternA \\($\timestep = 1000$)
    \end{minipage} &
    \begin{minipage}{0.15\linewidth}
    \centering
    \includegraphics[width=1\linewidth]{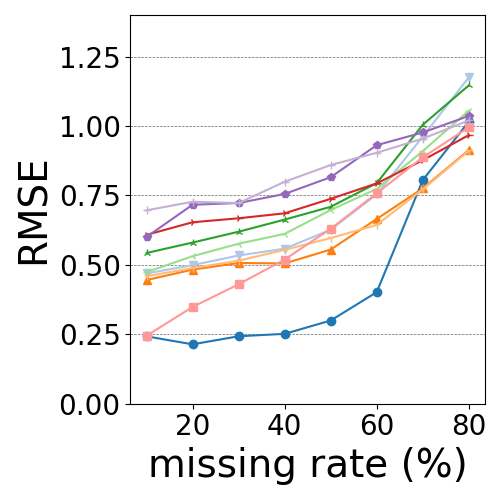} \\
    \vspace{-.5em}
    (b) \patternB \\($\timestep = 1000$)
    \end{minipage} &
    \begin{minipage}{0.15\linewidth}
    \centering
    \includegraphics[width=1\linewidth]{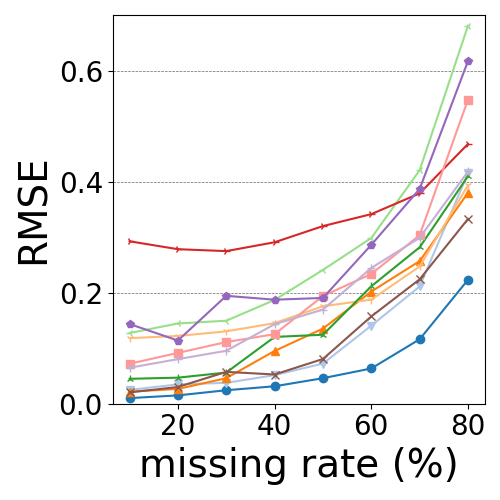} \\
    \vspace{-.5em}
    (c) Run \\($\timestep = 145$)
    \end{minipage} &
    \begin{minipage}{0.15\linewidth}
    \centering
    \includegraphics[width=1\linewidth]{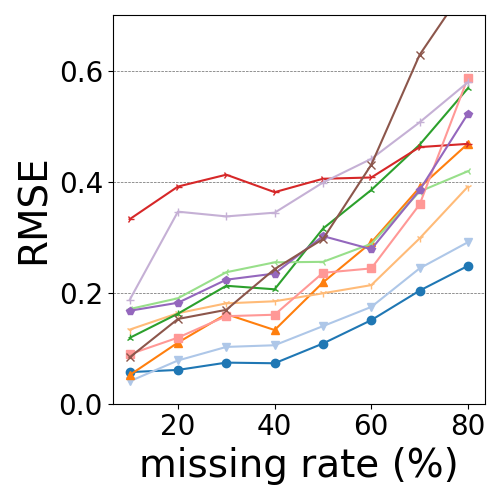} \\
    \vspace{-.5em}
    (d) Bouncy walk \\($\timestep = 644$)
    \end{minipage} &
    \begin{minipage}{0.15\linewidth}
    \centering
    \includegraphics[width=1\linewidth]{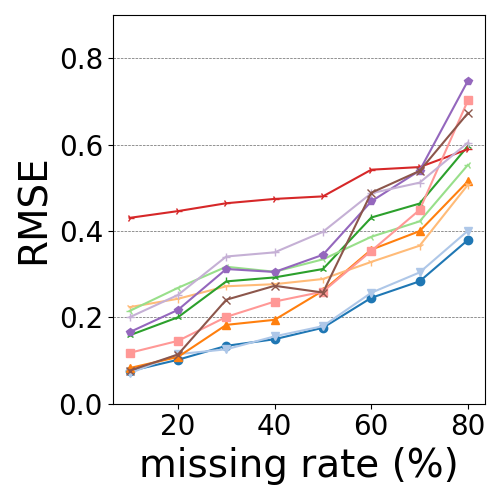} \\
    \vspace{-.5em}
    (e) Swing shoulder \\($\timestep = 804$)
    \end{minipage} &
    \begin{minipage}{0.15\linewidth}
    \centering
    \includegraphics[width=1\linewidth]{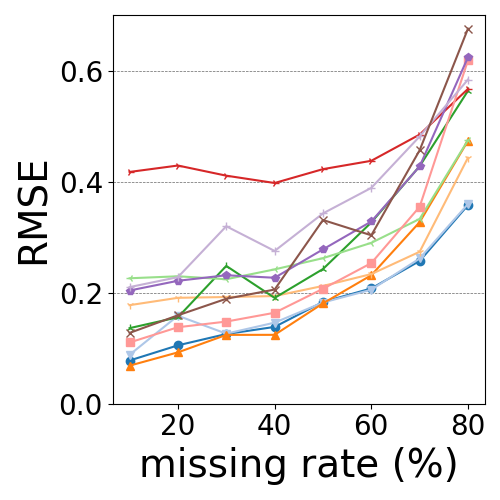} \\
    \vspace{-.5em}
    (f) Walk slow \\($\timestep = 1223$)
    \end{minipage} \\
    \end{tabular}
    \end{minipage}

    \begin{minipage}{1.00\linewidth}
    \centering
    \begin{tabular}{cccccc}
    \hspace{-1.000em}
    \begin{minipage}{0.15\linewidth}
    \centering
    \includegraphics[width=1\linewidth]{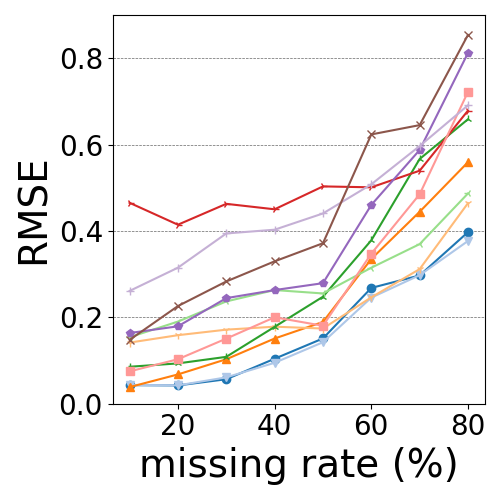} \\
    \vspace{-.5em}
    (g) Mawashigeri \\($\timestep = 1472$)
    \end{minipage} &
    \begin{minipage}{0.15\linewidth}
    \centering
    \includegraphics[width=1\linewidth]{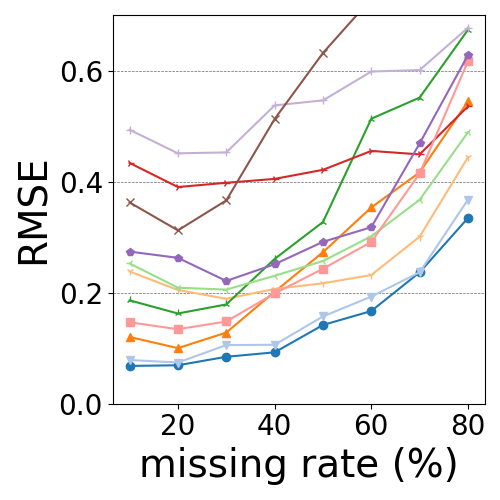} \\
    \vspace{-.5em}
    (h) Jump distance \\($\timestep = 547$)
    \end{minipage} &
    \begin{minipage}{0.15\linewidth}
    \centering
    \includegraphics[width=1\linewidth]{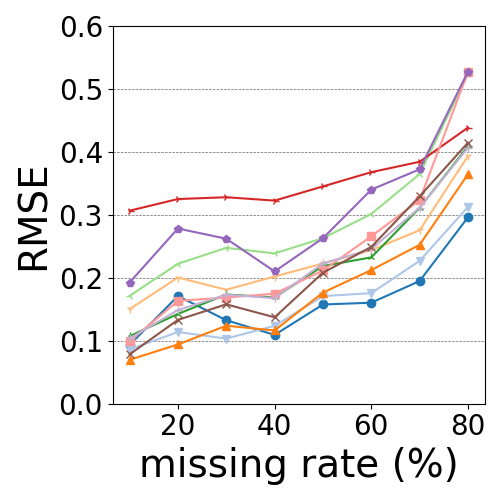} \\
    \vspace{-.5em}
    (i) Wave hello \\($\timestep = 299$)
    \end{minipage} &
    \begin{minipage}{0.15\linewidth}
    \centering
    \includegraphics[width=1\linewidth]{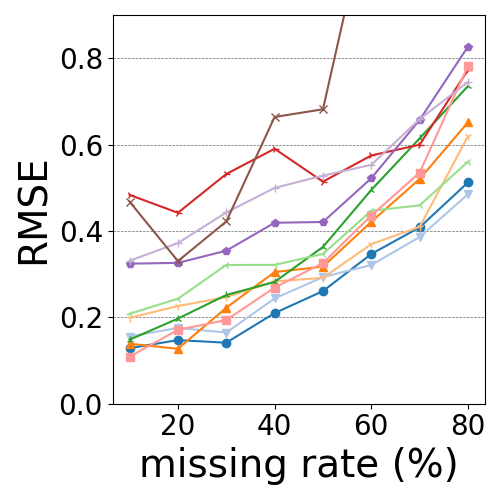} \\
    \vspace{-.5em}
    (j) Football throw \\($\timestep = 1091$) 
    \end{minipage} &
    \begin{minipage}{0.15\linewidth}
    \centering
    \includegraphics[width=1\linewidth]{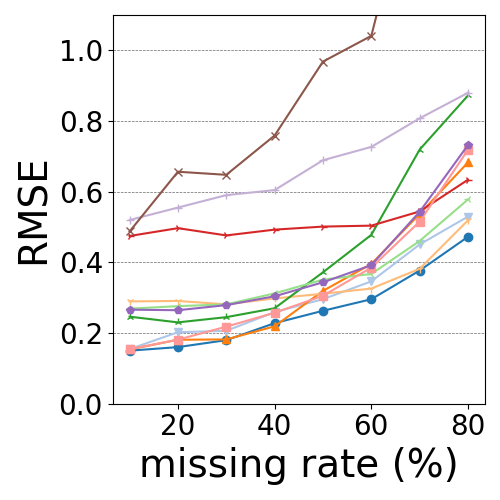} \\
    \vspace{-.5em}
    (k) Boxing \\($\timestep = 773$)
    \end{minipage} &
    \begin{minipage}{0.15\linewidth}
    \centering
    \includegraphics[width=1\linewidth]{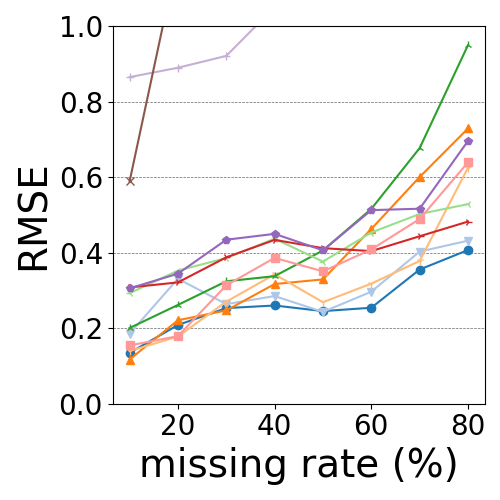} \\
    \vspace{-.5em}
    (l) Motes \\($\timestep = 336$)
    \end{minipage} \\
    \end{tabular}
    \end{minipage}

    \vspace{-1em}
    \caption{
    RMSE of (a), (b) \sync ($\ndim = 50$), (c) $\sim$ (k) \mocap ($\ndim = 123$) and (l) \motes ($\ndim = 54$) datasets.
    }
    \label{fig:rmse}
    \vspace{-1em}
\end{figure*}
\begin{figure}[t]
    \centering
    \begin{minipage}{1\columnwidth}
    \centering
    \includegraphics[width=1\linewidth]{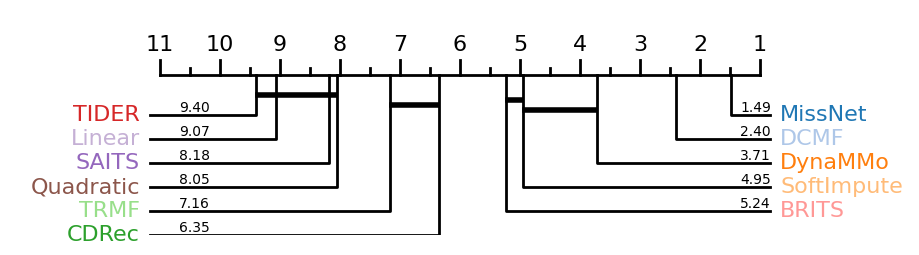} \\
    \end{minipage}
    \vspace{-1em}
    \caption{
        Critical difference diagram of real-world datasets.
        }
    \label{fig:cd}
    \vspace{-1em}
\end{figure}

In this section, we empirically evaluate our approach against state-of-the-art baselines on 12 datasets.
We present experimental results for the following questions:
\\ \textbf{Q1. Effectiveness}: How accurate is the proposed \method for recovering missing values?
\\ \textbf{Q2. Scalability}: How does the proposed algorithm scale?
\\ \textbf{Q3. Interpretability}: How can \method help us understand the data?

\subsection{Experimental setup}

\subsubsection{Datasets}
We use the following datasets.

\myparaitemize{\sync}
We generate two types of synthetic data, \patternA and \patternB, five times each,
by defining $\latentfactor$ and $\objectmatrixset$.
We set $\timestep = 1000, \ndim = 50$ and $\latentdim = 10$ (\apdx{\ref{apd:synthetic}}).
\patternA has one regime ($\nstate=1$), and in \patternB ($\nstate=2$), two regimes switch at every $200$ timesteps.

\myparaitemize{\mocap}
This dataset contains nine types of full body motions of \mocap database
\footnote{\url{http://mocap.cs.cmu.edu}}.
Each motion measures the positions of $41$ bones in the human body, resulting in a total of $123$ features  (X, Y, and Z coordinates).

\myparaitemize{\motes}
This dataset consists of temperature measurements from the $54$ sensors deployed in the Intel Berkeley Research Lab
\footnote{\url{https://db.csail.mit.edu/labdata/labdata.html}}.
We use hourly data for the first two weeks (03-01 $\sim$ 03-14).
Originally, $9.6\%$ of the data is missing, including a blackout from 03-10 to 03-11 where all the values are missing.

\subsubsection{Data preprocessing}
We generate a synthetic missing block at a length of $0 \sim 5\%$ of the data length and place it randomly until the total missing rate reaches $\{10, 20, \dots 80\% \}$.
Thus, a missing block can be longer than $0.05\timestep$ when it overlaps.
An additional $10\%$ of missing values are added for hyperparameter tuning.
Each dataset feature is normalized independently using a z-score so that each dataset has a zero mean and a unit variance.

\subsubsection{Comparison methods}
We compare our method with state-of-the-art imputation methods ranging from
classical baselines (Linear and Quadratic),
MF-based methods (\softimpute, \cdrec and \trmf),
SSM-based methods (\dynammo and \dcmf),
to DL-based methods (\brits, \saits and \tider).
\begin{itemize}
    \item Linear/Quadratic \footnote{\url{https://pandas.pydata.org/docs/reference/api/pandas.DataFrame.interpolate.html}}
    use linear/quadratic equations to interpolate missing values.
    \item \softimpute~\cite{softimpute} first fills in missing values with zero, then alternates between recovering the missing values and updating the SVD using the recovered matrix.
    \item \cdrec~\cite{cdrec} is a memory-efficient algorithm that iterates centroid decomposition (CD) and missing value imputation until they converge.
    \item \trmf~\cite{trmf} is based on MF that imposes temporal dependency among the data points with the AR model.
    \item \dynammo~\cite{dynammo} first fills in missing values using linear interpolation and then uses the EM algorithm to iteratively recover missing values and update the LDS model.
    \item \dcmf~\cite{dcmf} adds a contextual constraint to SSM and captures inter-correlation by a predefined network.
    As suggested in the original paper, we give the cosine similarity between each pair of time series calculated after linear interpolation as a predefined network.
    This method is similar to \method if we set $\nstate = 1$, employ a predefined network that is fixed throughout the algorithm, and eliminate the effect of regime-switching and network inference models from \method.
    \item \brits~\cite{brits} imputes missing values according to hidden states from bidirectional RNN.
    \item \saits~\cite{saits} is a self-attention-based model that jointly optimizes imputation and reconstruction to perform the missing value imputation of multivariate time series.
    \item \tider~\cite{tider} learns disentangled seasonal and trend representations by employing a neural network combined with MF.
\end{itemize}

\subsubsection{Hyperparameter setting}
For \method, we use the latent dimensions of $10,30$ and $15$ for \sync, \mocap and \motes, respectively,
and we set $\sparseparam = 1.0, \contextparam=0.5$ for all datasets.
We set the correct number of regimes on \sync datasets; we vary $\nstate = \{1,2,3\}$ for other datasets.
We list the detailed hyperparameter settings for the baselines in \apdx{\ref{apd:hyperparam}}.

\subsubsection{Evaluation metric}
To evaluate the effectiveness, we use Root Mean Square Error (RMSE) of the observed time series~\cite{dcmf}.

\subsection{Results}

\subsubsection{Q1. Effectiveness}\label{sec:q1}
We show the effectiveness of \method over baselines in missing value imputation.

\myparaitemize{\sync}
\fig{\ref{fig:rmse}} (a) and (b) show the results obtained with \sync datasets.
SSM and MF-based methods perform worse with \patternB than with \patternA due to the increased complexity of data.
DL-based methods, especially \brits, are less affected thanks to their high modeling power.
\method significantly outperforms \dcmf for \patternB although it produces similar results for \patternA.
This is because \dcmf fails to capture inter-correlation with \patternB since it can only use one predefined network and cannot afford a change of network.
Meanwhile, \method can capture the inter-correlation for two different regimes thanks to our regime-switching model.
However, \method fails to discover the correct transition when the missing rate exceeds $70\%$, and RMSE becomes similar to \dcmf.

\myparaitemize{\mocap and \motes}
The results for \mocap and \motes datasets are shown in \fig{\ref{fig:rmse}} (c) $\sim$ (l).
We can see that \method and \dcmf constantly outperform other baselines thanks to their ability to exploit inter-correlation explicitly.

\fig{\ref{fig:cd}} shows the corresponding critical difference diagram for all missing rates based on the Wilcoxon-Holm method~\cite{ismail2019deep}, where methods that are not connected by a bold line are significantly different in average rank.
This confirms that \method significantly outperforms other methods, including \dcmf, in average rank.
Our algorithm for repeatedly inferring networks and the use of \lonenorm enables the inference of adequate networks for imputation, contributing to better results than \dcmf,
which uses cosine similarity as a predefined network that may contain spurious correlations in the presence of missing values.
Note that \method and \dcmf exhibit only minor differences when the missing rate is low ($10\%$) because a plausible predefined network can be calculated from observed data.

Classical Linear and Quadratic baselines are unsuitable for imputing missing blocks since they impute missing values mostly from neighboring observed points and cannot capture temporal patterns when there are large gaps.
DL-based methods lack sufficient training data and are not suitable for the data we use here, making them perform particularly poorly at a high missing rate, as also noted in~\cite{missingvldb}.
MF-based methods, \softimpute and \cdrec, have a higher RMSE than SSM-based methods since they do not model the temporal dynamics of the data.
\trmf utilizes temporal dependency with the AR model, however, it can only capture certain lags specified on the hyperparameter of the AR model.
SSM-based methods are superior in imputation to other groups owing to their ability to capture temporal dependency in latent space.
\dynammo implicitly captures inter-correlation in latent space.
Therefore, it is no match for \method or \dcmf.

\begin{figure}[t]
    \centering
    \begin{tabular}{cc}

    \hspace{-2em}
    \begin{minipage}{0.7\columnwidth}
    \centering
    \includegraphics[width=0.9\linewidth]{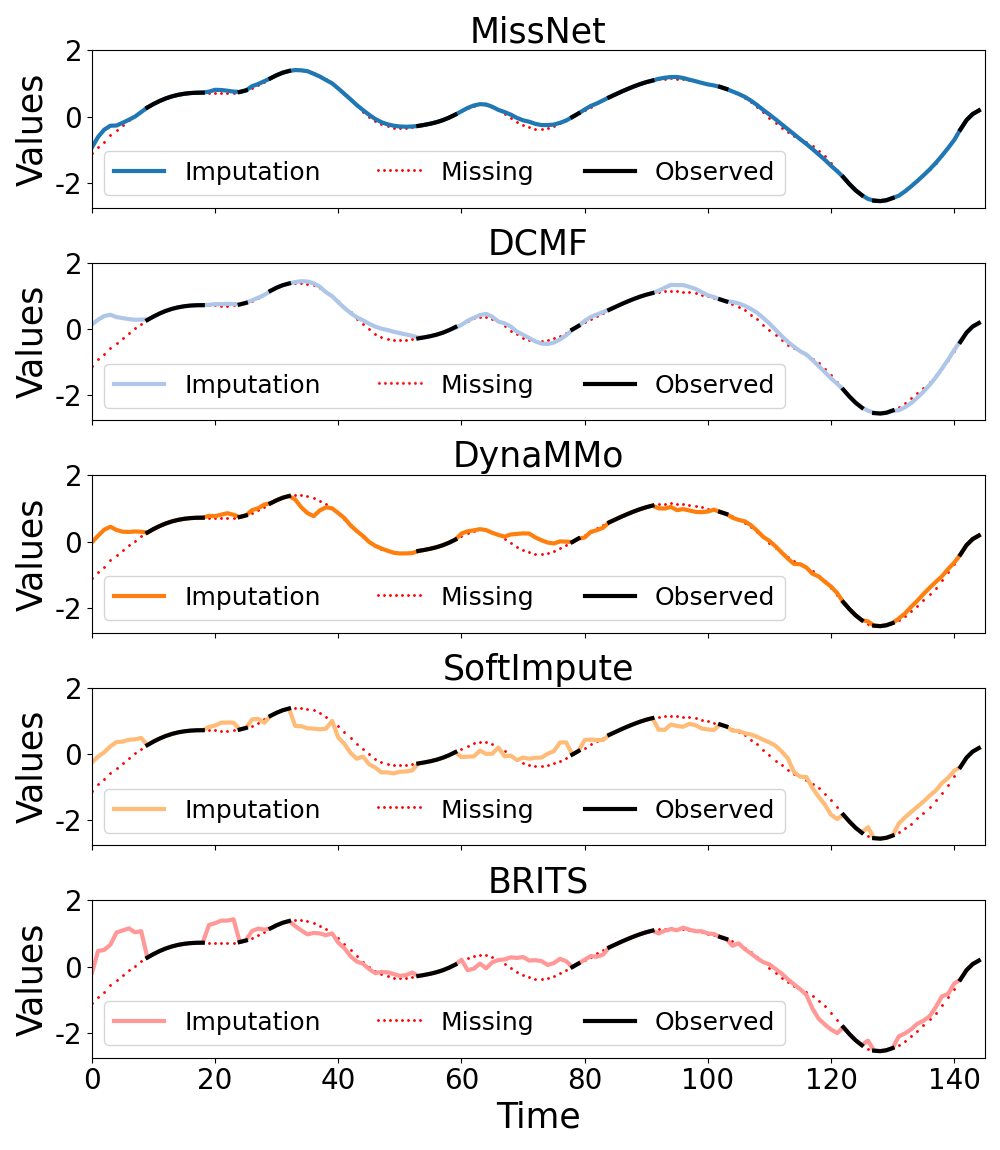} \\
    (a) Imputation results at RKNE
    \vspace{0em}
    \end{minipage} &
    
    \hspace{-4em}
    \begin{minipage}{0.5\columnwidth}
    \centering
    \vspace{1em}
    \includegraphics[width=0.6\linewidth]{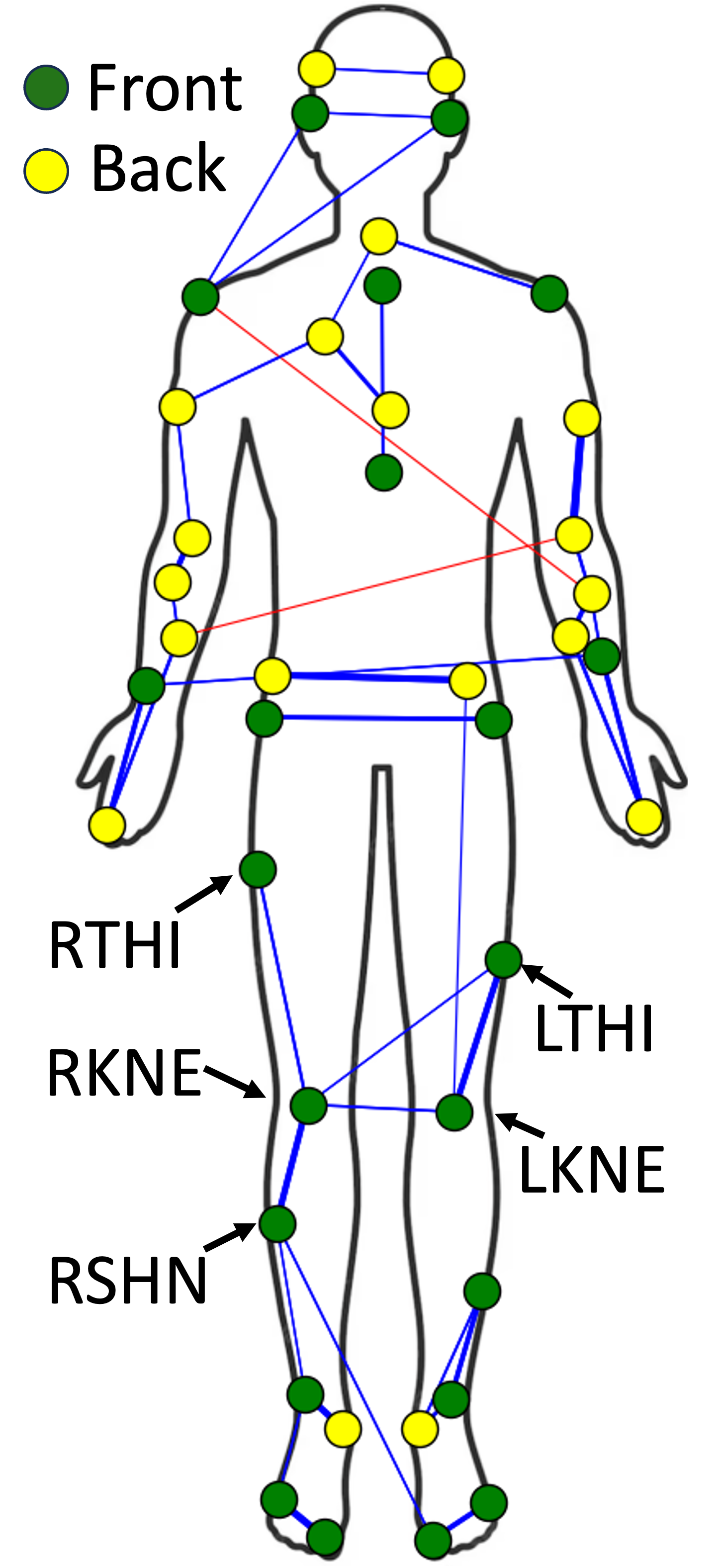} \\
    \vspace{1em}
    (b) Y-network of \method 
    \end{minipage} \\
    
    \end{tabular}
    \vspace{-1em}
    \caption{
        Case study on \mocap Run dataset.
        }
    \label{fig:comparison}
    \vspace{-1em}
\end{figure}

\fig{\ref{fig:comparison}} demonstrates the results for the \mocap Run dataset (missing rate $= 60 \%$).
We compare the imputation result for the sensor at RKNE provided by the top five methods in terms of average rank, including \method, in \fig{\ref{fig:comparison}} (a).
\brits and \softimpute fail to capture the dynamics of time series while providing a good fit to observed values.
The imputation of \dynammo is smooth, but some parts are imprecise since it cannot explicitly exploit inter-correlation.
\method and \dcmf can effectively exploit other observed features associated with RKNE, thereby accurately imputing missing values where other methods fail (e.g., $T=20\sim40, 60\sim80$).
\fig{\ref{fig:comparison}} (b) shows the sensor network of Y-coordinate values obtained by \method plotted on the human body, where a green/yellow dot (node) indicates a sensor placed on the front/back of the body and the thickness and color (blue/red) of the edges are the value and sign (positive/negative) of partial correlations, respectively.
We can see that the sensors located close together have edges, meaning they are conditionally dependent given all other sensor values.
For example, the sensor at RKNE has edges between RTHI, RSHN, LKNE, and LTHI.
They are located to RKNE nearby and show similar dynamics, thus it is reasonable to consider that they are connected.
Since \method can infer such a meaningful network from partially observed data, the imputation of \method is more accurate than that of \dcmf.

\begin{figure}[t]
    \centering
    \begin{minipage}{1\columnwidth}
    \centering
    \begin{tabular}{ccc}
    
    \begin{minipage}{0.3\columnwidth}
    \centering
    \includegraphics[width=1\linewidth]{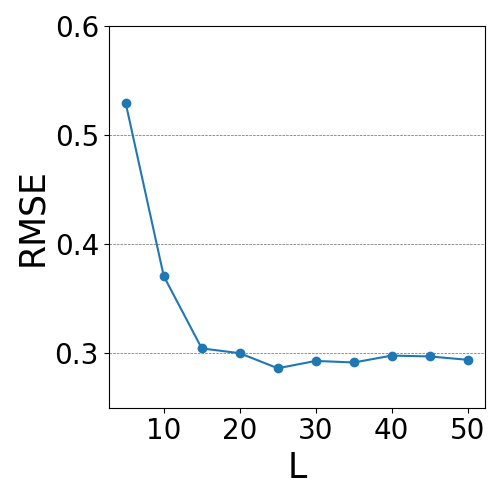} \\
    \vspace{-0.5em}
    (a) Impact of $\latentdim$
    \end{minipage} &
    \begin{minipage}{0.3\columnwidth}
    \centering
    \includegraphics[width=1\linewidth]{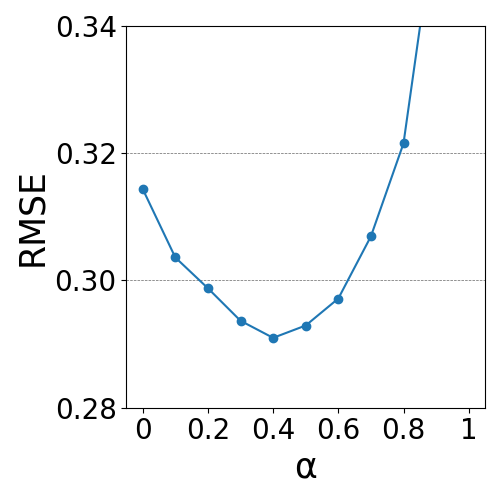} \\
    \vspace{-0.5em}
    (b) Impact of $\contextparam$
    \end{minipage} &
    \begin{minipage}{0.3\columnwidth}
    \centering
    \includegraphics[width=1\linewidth]{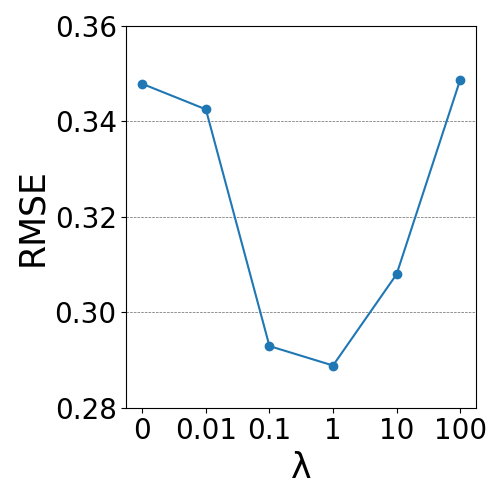} \\
    \vspace{-0.5em}
    (c) Impact of $\sparseparam$
    \end{minipage} \\
    
    \vspace{-2em}
    \end{tabular}
    \end{minipage}
    \caption{
    Hyperparameter sensitivity results.
    }
    \label{fig:param}
    \vspace{-1em}
\end{figure}
\subsubsection{Hyperparameter sensitivity}
We take the \motes dataset and show the impact of hyperparameters:
the latent dimension $\latentdim$, the network parameter $\contextparam$, and the sparse parameter $\sparseparam$.
We show the mean RMSE of all missing rates.

\myparaitemize{Latent dimension}
\fig{\ref{fig:param}} (a) shows the impact of $\latentdim$.
As $\latentdim$ becomes larger, the model's fitting against the observed data increases.
As we can see, the RMSE is constantly decreasing as $\latentdim$ increases and stabilizes after $15$.
This shows that \method does not overfit the observed data even for a large $\latentdim$.

\myparaitemize{Network parameter}
$\contextparam$ determines the contributions of inter-correlation and temporal dependency to learning $\objectmatrixset$.
If $\contextparam=0$, the contextual matrix $\contextualmatrixset$ is ignored.
If $\contextparam=1$, only $\contextualmatrixset$ is considered for learning $\objectmatrixset$.
\fig{\ref{fig:param}} (b) shows the results of varying $\contextparam$ and they are robust except when $\contextparam=1$ (RMSE $=0.76$).
We can see that $\contextparam = 0.4$ shows the best result, indicating that both temporal dependency and inter-correlation are important for precise imputation.

\myparaitemize{Sparse parameter}
$\sparseparam$ controls the sparsity of the networks $\invcovmatrixset$ through \lonenorm.
The bigger $\sparseparam$ becomes, the more sparse the networks become, resulting in \method considering only strong interplay.
By contrast, when $\sparseparam$ is small, \method considers insignificant interplays.
\fig{\ref{fig:param}} (c) shows the impact of $\sparseparam$.
We can see that the sparsity of the networks affects the accuracy, and the best $\sparseparam$ exists between $0.1$ and $10$.
Thus, the \lonenorm constraint helps \method to exploit important relationships.

\begin{figure}[t]
    \centering
    \begin{minipage}{1\columnwidth}
    \centering
    \includegraphics[width=1\linewidth]{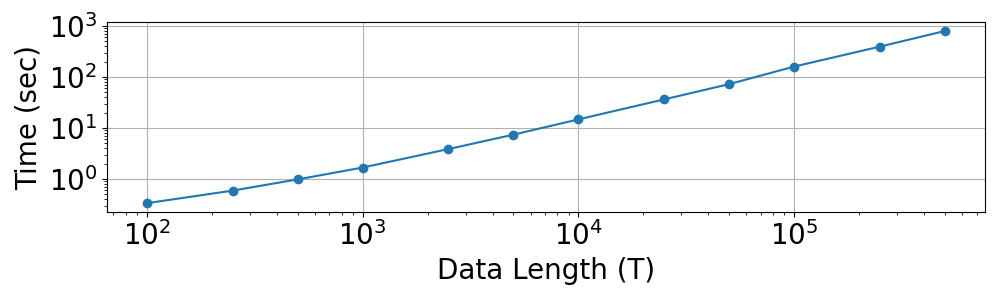} \\
    \end{minipage}
    \vspace{-1.5em}
    \caption{
    Scalability of \method.
    }
    \label{fig:scale}
    \vspace{-1em}
\end{figure}
\subsubsection{Q2. Scalability}
We test the scalability of the \method algorithm by changing the number of the data length ($\timestep$) in \patternA.
\fig{\ref{fig:scale}} shows the computation time for one iteration plotted with the data length.
As it shows, our proposed \method algorithm scales linearly with regard to the data length $\timestep$.

\begin{figure}[t]
    \centering
    \begin{minipage}{1\columnwidth}
    \centering
    \includegraphics[width=1\linewidth]{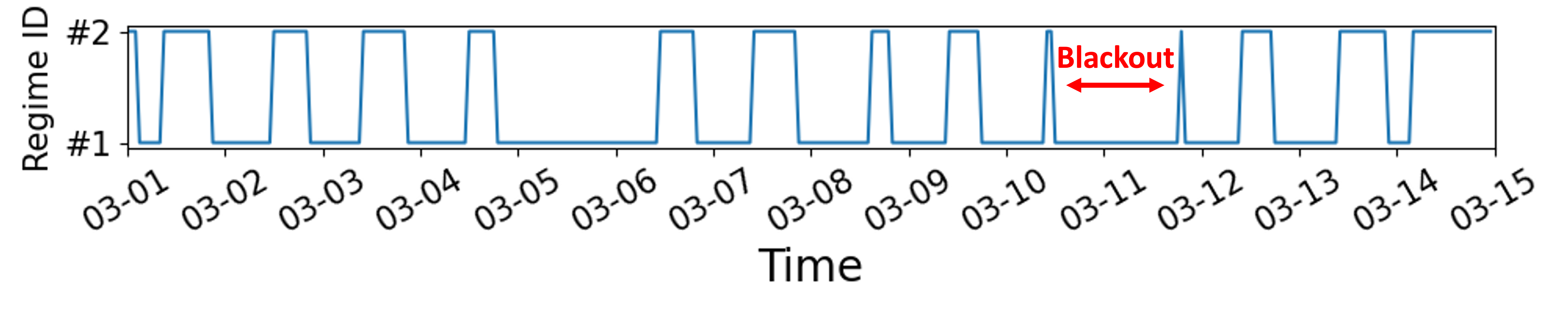} \\
    \vspace{-.5em}
    (a) Results of regime assignments
    \end{minipage} \\
    
    \vspace{0.5em}
    
    \begin{tabular}{cc}
    \begin{minipage}{0.47\columnwidth}
    \centering
    \includegraphics[width=1\linewidth]{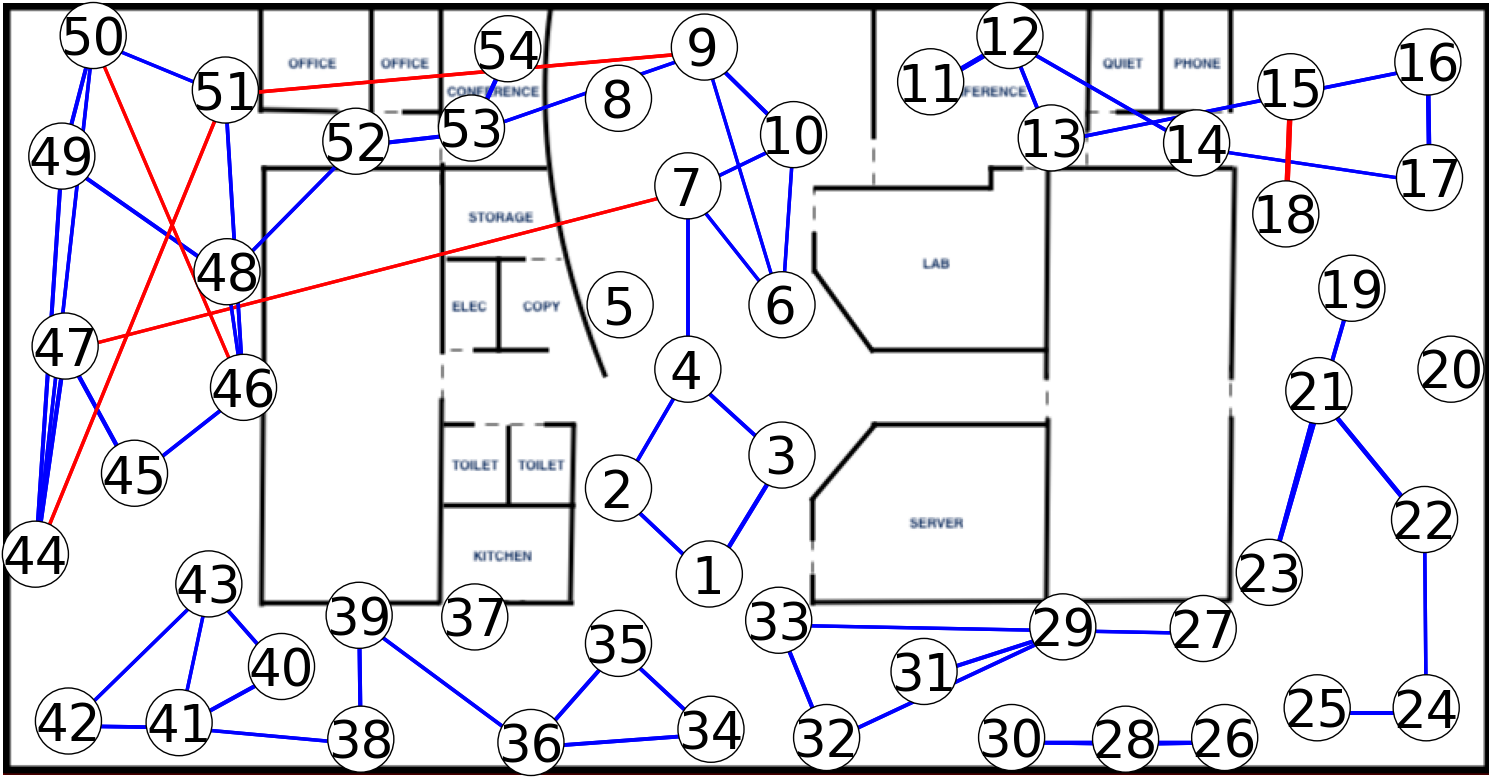} \\
    (b) Regime $\#1$, $\#$ of edges: 116 
    \end{minipage} &
    \begin{minipage}{0.47\columnwidth}
    \centering
    \includegraphics[width=1\linewidth]{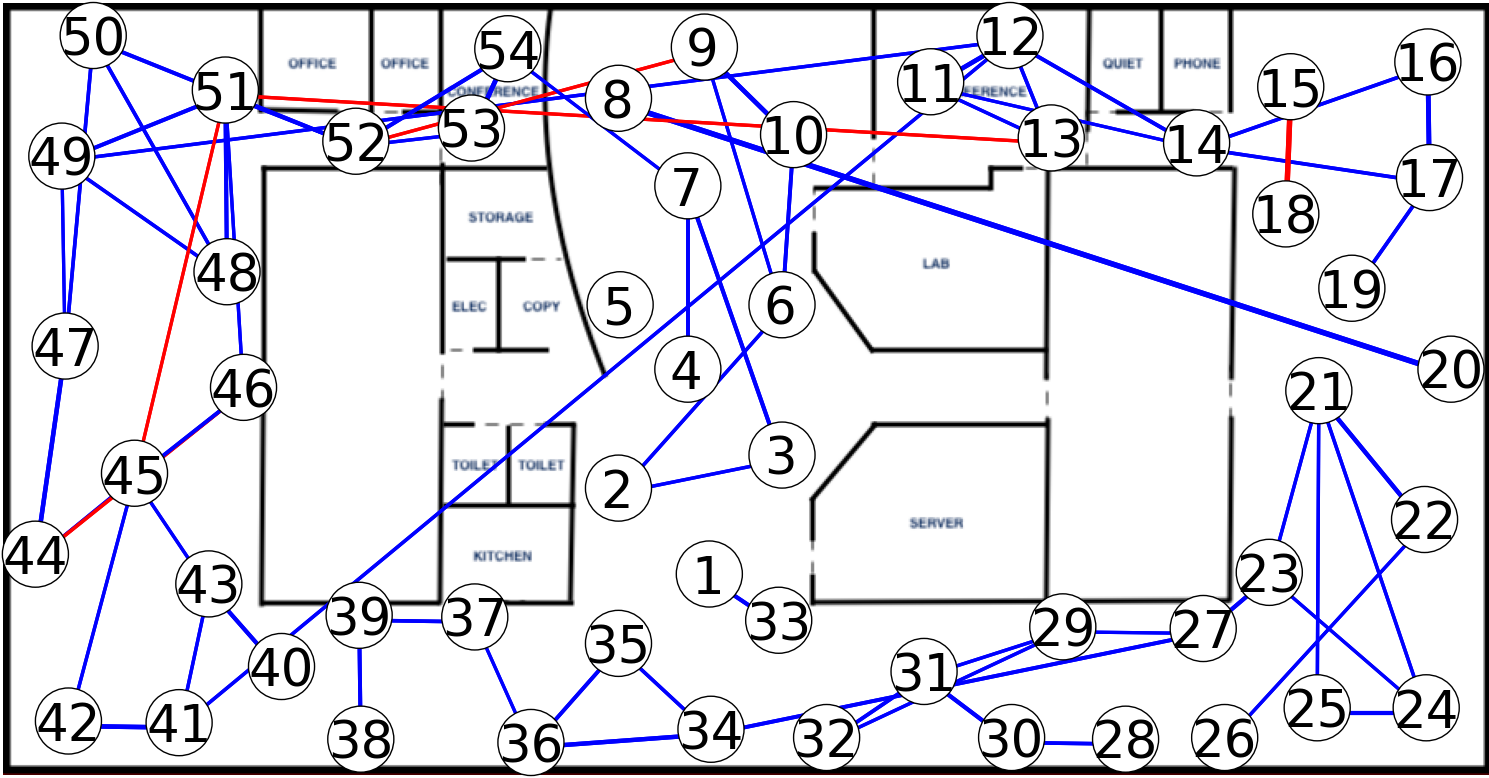} \\
    (c) Regime $\#2$, $\#$ of edges: 134 
    \end{minipage} \\
    \end{tabular}

    \vspace{-1em}
    \caption{
    Case study on \motes dataset.
    }
    \label{fig:motes}
    \vspace{-1em}
\end{figure}
\subsubsection{Q3. Interpretability}
We demonstrate how \method helps us understand data.
We have shown an example with the \mocap Run dataset in \fig{\ref{fig:comparison}} (b) where \method provides an interpretable network.
Here, we demonstrate the results on the \motes dataset (missing rate $= 30 \%$) of \method ($\nstate = 2$).
\fig{\ref{fig:motes}} (a) shows the regime assignments $\switchmatrix$, and \method mostly assigns night hours to regime $\#1$ and working hours (about $9$ am. to $10$ pm.) to regime $\#2$,
suggesting that they have different networks.
\fig{\ref{fig:motes}} (b) and (c) show the networks for regimes $\#1$ and $\#2$ obtained by \method plotted on the building layout.
The sensor numbers in the figure are plotted on the actual sensor deployments.
As we can see, the two regimes have different networks, and a common feature is that the neighboring sensors tend to form edges, which aligns with our expectations,
considering that the sensors measure temperature and, thus, neighboring sensors correlate.
The network of regime $\#2$ has more edges than that of $\#1$, and the edges are $1.2$ times longer on average, which might be caused by air convection due to the movement of people during working hours.
Consequently, \method provides regime assignments and sparse networks, which help us understand how data can be separated and important relationships for imputation.

\section{Conclusion}
    \label{060conclusions}
    In this paper, we proposed an effective missing value imputation method for multivariate time series, namely \method,
which captures temporal dependency based on latent space and inter-correlation by the inferred networks while discovering regimes.
Our proposed method has the following properties:
(a) \textit{Effective}: it outperforms the state-of-the-art algorithms for multivariate time series imputation.
(b) \textit{Scalable}: the computation time of \method scales linearly with regard to the length of the data.
(c) \textit{Interpretable}: it provides sparse networks and regime assignments, which help us understand the important relationships for imputation visually.
Our extensive experiments demonstrated the above properties of \method.

\begin{acks}
The authors would like to thank the anonymous referees for their valuable comments and helpful suggestions.
This work was supported by
JSPS KAKENHI Grant-in-Aid for Scientific Research Number
JP21H03446,     
JP22K17896,    
NICT JPJ012368C03501, 
JST-AIP JPMJCR21U4, 
JST CREST JPMJCR23M3.   
\end{acks}

\bibliographystyle{ACM-Reference-Format}
\balance
\bibliography{ref_missing}

\appendix
\label{100appendix}
\section{Proposed model}

\subsection{Symbols} \label{apd:symbol}
\tabl{\ref{table:symbol}} lists the main symbols we use throughout this paper.

\subsection{Complexity analysis} \label{apd:proof}
\myparaitemize{Proof of Lemma \ref{lemma:time}}
\begin{proof}
    The overall time complexity is composed of four parts by taking the most time-consuming part of equations for each iteration considering $\ndim > \ndim_{t}, \latentdim$:
    the complexity for the inference of $\latentfactor$ and $\switchmatrix$ is
    $O(\nstate^{2}\sum_{t=1}^\timestep(\latentdim^{3} + \latentdim^{2}\ndim_{t} + \latentdim\ndim_{t}^{2} + \ndim_{t}^{3}))$
    related to \eq{\ref{eq:forward}} and \eq{\ref{eq:partialcost}};
    the inference of $\objectfactorset$ is
    $O(\nstate\latentdim^{2}\ndim^{2})$
    (\eq{\ref{eq:objectmatrix}});
    M step is
    $O(\nstate\timestep\latentdim^{2}\ndim)$
    related to the calculation of $\objectmatrixset$ (\eq{\ref{eq:paramobject}});
    and the update of $\invcovmatrixset$ is
    $O(\nstate\ndim^{3})$
    (\eq{\ref{eq:network}}).
    Thus, the overall time complexity is
    $O( \#iter \cdot ( \nstate^{2}\sum_{t=1}^\timestep(\latentdim^{3} + \latentdim^{2}\ndim_{t} + \latentdim\ndim_{t}^{2} + \ndim_{t}^{3}) + \nstate\latentdim^{2}\ndim^{2} + \nstate\timestep\latentdim^{2}\ndim + \nstate\ndim^{3} ) )$.
\end{proof}

\myparaitemize{Proof of Lemma \ref{lemma:space}}
\begin{proof}
    The space complexity is composed of three parts:
    storing input dataset $\mts$ is $O( \timestep\ndim)$;
    intermediate values in E step are $O(\nstate^{2}\timestep\latentdim^{2} + \nstate\latentdim^{2}\ndim)$;
    and storing parameter set is $O(\nstate\ndim^{2})$.
    Thus, the overall space complexity is
    $O( \timestep\ndim + \nstate^{2}\timestep\latentdim^{2} + \nstate\latentdim^{2}\ndim + \nstate\ndim^{2} )$.
\end{proof}

\begin{table}[ht]
    \centering
    \small
    \caption{Symbols and definitions.}
    \label{table:symbol}
    \begin{tabular}{l|p{6cm}}
        \hline \hline
        Symbol & Definition \\
        \hline
        $\mts$ & Multivariate time series $\mts = \{ \mtsvector_1,\mtsvector_2,\dots,\mtsvector_\timestep \} \in \realnumber^{\ndim \times \timestep}$ \\
        $\indicatormatrix$ & Indicator matrix $\indicatormatrix \in \realnumber^{\ndim \times \timestep}$ \\
        $\givenmts$ & Partially observed multivariate time series $\givenmts = \indicatormatrix \hadamard \mts$ \\
        $\impmts$ & Inputed multivariate time series \\
        \hline
        $\latentfactor$ & Time series latent states $\latentfactor \in \realnumber^{\latentdim \times \timestep}$ \\
        $\objectfactorN{k}$ & Contextual latent factor of \thN{k}-regime $\objectfactorN{k} \in \realnumber^{\latentdim \times \ndim}$ \\
        $\contextualmatrixN{k}$ & Contextual matrix of \thN{k}-regime $\contextualmatrixN{k} \in \realnumber^{\ndim \times \ndim}$ \\
        $\transitionmatrix$ & Transition matrix $\transitionmatrix \in \realnumber^{\latentdim \times \latentdim}$ \\
        $\objectmatrixN{k}$ & Observation matrix of \thN{k}-regime $\objectmatrixN{k} \in \realnumber^{\ndim \times \latentdim}$ \\
        $\switchmatrix$ & Regime assignments $\switchmatrix \in \realnumber^{\nstate \times \timestep}$ \\
        $\meanvectorN{k}$ &  Mean vector of \thN{k}-regime $\meanvectorN{k} \in \realnumber^{\ndim}$ \\
        $\invcovmatrixN{k}$ & Inverse covariance matrix (i.e., network) of \thN{k}-regime $\invcovmatrixN{k} \in \realnumber^{\ndim \times \ndim}$ \\
        \hline
        $\ndim$ & Number of features\\
        $\timestep$ & Number of timesteps\\
        $\latentdim$ & Number of latent dimensions\\
        $\nstate$ & Number of regimes\\
        $\contextparam$ & Trade-off between temporal dependency and inter-correlation\\
        $\sparseparam$ & Parameter for \lonenorm that regulates network sparsity\\
        \hline
    \end{tabular}
\end{table}

\subsection{Updating parameters} \label{apd:algparam}
The parameters are updated as follows:
\begin{align}
    \transitionmatrix^{new} &= \bigg( \sum_{t=2}^{\timestep}\expectation{\latentfactorvector_{t} \latentfactorvector'_{t-1}} \bigg) \bigg( \sum_{t=2}^{\timestep}\expectation{\latentfactorvector_{t-1} \latentfactorvector'_{t-1}} \bigg)^{-1} , \nonumber \\
    \latentfactorvector_{0}^{new} &= \expectation{\latentfactorvector_{1}} , \nonumber \\
    \sgmZero^{new} &= \expectation{\latentfactorvector_{1} \latentfactorvector'_{1}} - \expectation{\latentfactorvector_{1}}\expectation{\latentfactorvector'_{1}} , \nonumber
\end{align}
\begin{align}
    (\sgmZ^{2})^{new} =& \frac{1}{(\timestep-1) \latentdim} \textrm{tr} 
    \bigg( \sum_{t=2}^{\timestep} \expectation{\latentfactorvector_{t}\latentfactorvector'_{t}}
    -\transitionmatrix\sum_{t=2}^{\timestep} \expectation{\latentfactorvector_{t}\latentfactorvector'_{t}} \bigg) , \nonumber \\
    (\sgmXN{k}^{2})^{new} =& \frac{1}{\sum_{t=1 (\switchvector_{t}=k)}^{\timestep} \sum_{i=1}^{\ndim} \indicatormatrix_{i,t} } \nonumber \\
    & \bigg[ \sum_{t=1 (\switchvector_{t}=k)}^{\timestep} tr \bigg( (\obsobjectmatrixN{k}_{t})^{new} \expectation{\latentfactorvector_{t} \latentfactorvector'_{t}}(\obsobjectmatrixN{k}_{t})^{new'} \bigg) \nonumber \\
    &+ \sum_{t=1 (\switchvector_{t}=k)}^{\timestep} \bigg( (\obsmtsvector_{t})'\obsmtsvector_{t} - 2(\obsmtsvector_{t})'( (\obsobjectmatrixN{k}_{t})^{new} \expectation{\latentfactorvector_{t}} ) \bigg) \bigg] , \nonumber \\
    (\sgmSN{k}^{2})^{new} =& \frac{1}{\ndim^2} \bigg[ \sum_{j=1}^{\ndim} \bigg( \contextualvector^{(k)'}_{j}\contextualvector^{(k)}_{j} - 2\contextualvector^{(k)'}_{j}( (\objectmatrixN{k})^{new} \expectation{\objectfactorvector^{(k)}_{j}}) \bigg) \nonumber \\
    &+ tr \bigg( (\objectmatrixN{k})^{new} (\sum_{j=1}^{\ndim} \expectation{\objectfactorvector^{(k)}_{j}\objectfactorvector^{(k)'}_{j}}) (\objectmatrix^{(k)new})' \bigg)  \bigg] , \nonumber \\
    (\sgmVN{k}^{2})^{new} =& \frac{1}{\ndim \latentdim} \sum_{j=1}^{\ndim} \textrm{tr} (\expectation{\objectfactorvectorN{k}_{j}\objectfactorvector^{(k)'}_{j}}) . \label{eq:paramlatent}
\end{align}

\section{Experiments}

\subsection{Synthetic data generation} \label{apd:synthetic}
We first generate a latent factor containing a linear trend, a sinusoidal seasonal pattern, and a noise,
$\latentfactor_{i,t} = \sin(2\pi \frac{\beta}{\timestep} t) + \gamma t + \eta$, s.t. $ 1 < \beta < 20, 0.3 < |\gamma| < 1, \eta \sim \gauss{0}{0.3} $, where $\latentfactor \in \realnumber^{\latentdim \times \timestep}$.
We then project the latent factor with object matrix $\mts = \objectmatrixN{k} \latentfactor$,
where $\mts \in \realnumber^{\ndim \times \timestep}$,
and $\objectmatrixN{k} \in \realnumber^{\ndim \times \latentdim}$ is a random graph created as follows~\cite{mohan14}:
\begin{enumerate}
    \setlength{\parskip}{0cm}
    \setlength{\itemsep}{0cm}
    \item Set $\objectmatrixN{k}$ equal to the adjacency matrix of an Erd\H{o}s-R\'{e}nyi directed random graph, where every edge has a $20\%$ chance of being selected.
    \item Set selected edge $\objectmatrixN{k}_{i,j}\sim$ Uniform$([-0.6,-0.3]\cup[0.3,0.6])$, where $\objectmatrixN{k}_{i,j}$ denotes the weight between variables $i$ and $j$.
\end{enumerate}
We set $\timestep = 1000, \ndim = 50, \latentdim = 10$.
We generate two types of synthetic data, \patternA and \patternB, five times each,
where $\nstate = 1, 2$, respectively.
In \patternB, the regime switches every $200$ timesteps.

\subsection{Hyperparameters} \label{apd:hyperparam}
We describe the hyperparameters of the baselines.
For \sync datasets, we give a latent dimension of $10$ for all baselines.
For a fair comparison, we set the latent dimension of the SSM-based methods at the same value as \method.
For the MF-based methods, we vary the latent dimension $\{ 3, 5, 10, 15, 20, 30, 40 \}$.
We vary the AR parameter for \trmf $\{ [1,2,3,4,5],[1,24] \}$.
To learn the DL-based methods, we add $10\%$ of the data as missing values for training the model.
We vary the window size $\{ 16, 32, \timestep \}$.
Other hyperparameters are the same as the original codes.

\section{Discussion}
While \method achieved superior performance against state-of-the-art baselines in missing value imputation, here, we mention two limitations of \method in terms of sparse network inference and data size.

As mentioned in \secton{\ref{sec:q1}}, \method fails to discover the correct
transition when the missing rate exceeds $70\%$.
However, we claim that \method failing to discover the correct transition when the missing rate exceeds $70\%$ is reasonable; rather, correctly discovering transition up to $60\%$ is valuable.
Several studies \cite{mmgl,kolar2012estimating,stadler2012missing} tackled the sparse network inference under the existence of missing values.
They aim to infer the correct network and, thus, only utilize the observed value for the network inference.
Since observing a complete pair at a high missing rate is rare, it is difficult to infer the correct network.
Therefore, the maximum missing rate in their experiments is $30\%$.
Although the experimental settings are different from ours, we can say that the task of sparse network inference in the presence of missing values itself is challenging.

As shown in the experiments, \method performs well even when a relatively small number of samples ($T$) and a large number of features ($N$) since \method is a parametric model and we assume the sparse networks to capture inter-correlation.
This cannot be achieved by DL models, which contain a massive number of parameters that require a large amount of $T$, especially when $N$ is large since all the relationships between features need to be learned.
However, unlike DL models, the increased number of samples may not greatly improve \method's performance as it has a much smaller number of parameters than DL models, even though switching sparse networks increases the model's flexibility.

\end{document}